%% file: main.tex
\documentclass[10pt,conference]{IEEEtran}
\usepackage{cite}
\usepackage{amsmath,amssymb,amsfonts}
\usepackage{algorithmic}
\usepackage{graphicx}
\usepackage{textcomp}
\usepackage[hyphens]{url}

\usepackage[normalem]{ulem}

\usepackage[dvipsnames]{xcolor}
\definecolor{light-blue}{HTML}{0095d9}
\definecolor{dark-orange}{HTML}{e17b34} 
\definecolor{dark-red}{HTML}{bf242a}  
\definecolor{dark-green}{HTML}{007b43} 
\definecolor{dark-purple}{HTML}{8d4bbb}  
\usepackage{algorithm2e}
\SetKwComment{Comment}{$\triangleright$\ }{}
\RestyleAlgo{ruled}
\SetKwInput{KwInput}{Input}              
\SetKwInput{KwOutput}{Output}
\usepackage{marvosym}
\usepackage{booktabs}

\usepackage{mathtools}
\usepackage{amsthm}
\usepackage{multirow}
\DeclareMathAlphabet\mathbfcal{OMS}{cmsy}{b}{n}
\usepackage{pifont}
\newtheorem{property}{Property}
\usepackage{subfiles}

\def\BibTeX{{\rm B\kern-.05em{\sc i\kern-.025em b}\kern-.08em
    T\kern-.1667em\lower.7ex\hbox{E}\kern-.125emX}}

\usepackage[bookmarks=true,breaklinks=true,colorlinks,citecolor=blue,linkcolor=blue,urlcolor=blue]{hyperref}
\usepackage{wrapfig}

\title{ViTALiTy: Unifying Low-rank and Sparse Approximation for \underline{Vi}sion \underline{T}ransformer \underline{A}cceleration with a \underline{Li}near \underline{T}a\underline{y}lor Attention} 

\author{
\IEEEauthorblockN{Jyotikrishna Dass*\IEEEauthorrefmark{2}, Shang Wu*\IEEEauthorrefmark{2}, Huihong Shi*\IEEEauthorrefmark{3}, Chaojian Li\IEEEauthorrefmark{3}, Zhifan Ye\IEEEauthorrefmark{2}, Zhongfeng Wang\IEEEauthorrefmark{4} and Yingyan Lin\IEEEauthorrefmark{3}}
\IEEEauthorblockA{\IEEEauthorrefmark{2}Rice University, Houston, TX \\ Email: \emph{\{jdass, sw99, zy50\}@rice.edu}}
\IEEEauthorblockA{\IEEEauthorrefmark{3}Georgia Institute of Technology, Atlanta, GA \\ Email: \emph{eiclab.gatech@gmail.com}, \emph{\{cli851, celine.lin\}@gatech.edu}}
\IEEEauthorblockA{\IEEEauthorrefmark{4}Nanjing University, Nanjing, Jiangsu \\ Email: \emph{zfwang@nju.edu.cn}}
}

\begin{document}
\maketitle

\def\thefootnote{}
\footnotetext{*The authors contributed equally to this work. JD proposed the core idea, algorithm design, experimental analysis, and led the paper writing, SW performed algorithmic implementation and software-related experiments, and HS led the hardware-related design, implementation, experiments and writing.}
\thispagestyle{plain}
\pagestyle{plain}

\begin{abstract}
Vision Transformer (ViT) has emerged as a competitive alternative to convolutional neural networks for various computer vision applications. Specifically, ViTs’ multi-head attention layers make it possible to embed information globally across the overall image. Nevertheless, computing and storing such attention matrices incurs a quadratic cost dependency on the number of patches, limiting its achievable efficiency and scalability and prohibiting more extensive real-world ViT applications on resource-constrained devices. Sparse attention has been shown to be a promising direction for improving hardware acceleration efficiency for NLP models. However, a systematic counterpart approach is still missing for accelerating ViT models. To close the above gap, we propose a first-of-its-kind algorithm-hardware codesigned framework, dubbed \textsc{ViTALiTy}, for boosting the inference efficiency of ViTs. Unlike sparsity-based Transformer accelerators for NLP, \textsc{ViTALiTy} unifies both low-rank and sparse components of the attention in ViTs. \underline{At the algorithm level}, we approximate the dot-product softmax operation via first-order Taylor attention with row-mean centering as the low-rank component to linearize the cost of attention blocks and further boost the accuracy by incorporating a sparsity-based regularization. \underline{At the hardware level}, we develop a dedicated accelerator to better leverage the resulting workload and pipeline from \textsc{ViTALiTy}'s linear Taylor attention which requires the execution of only the low-rank component, to further boost the hardware efficiency. Extensive experiments and ablation studies validate that \textsc{ViTALiTy} offers boosted end-to-end efficiency (e.g., $3\times$ faster and $3\times$ energy-efficient) under comparable accuracy, with respect to the state-of-the-art solution. 
\end{abstract}

\section{Introduction}
Vision Transformers (ViT) are gaining increasing popularity with their state-of-the-art performance in various computer vision tasks~\cite{vit, pmlr-v139-touvron21a,heo2021rethinking,Graham_2021_ICCV,liu2021Swin,wu2021cvt}. Compared to Convolutional Neural Networks (CNNs) which exploit local information through convolutional layers, ViT uses multi-head attention (MHA) modules to capture global information and long-range interactions, showing superior accuracy against CNNs \cite{vit}. Nevertheless, computing and storing such attention matrices incurs a quadratic computational and memory cost dependency on the number of patches (input resolution).

\begin{figure}[ht]
    \centering
    \vspace{-0.5em}
    \includegraphics[width=0.9\linewidth]{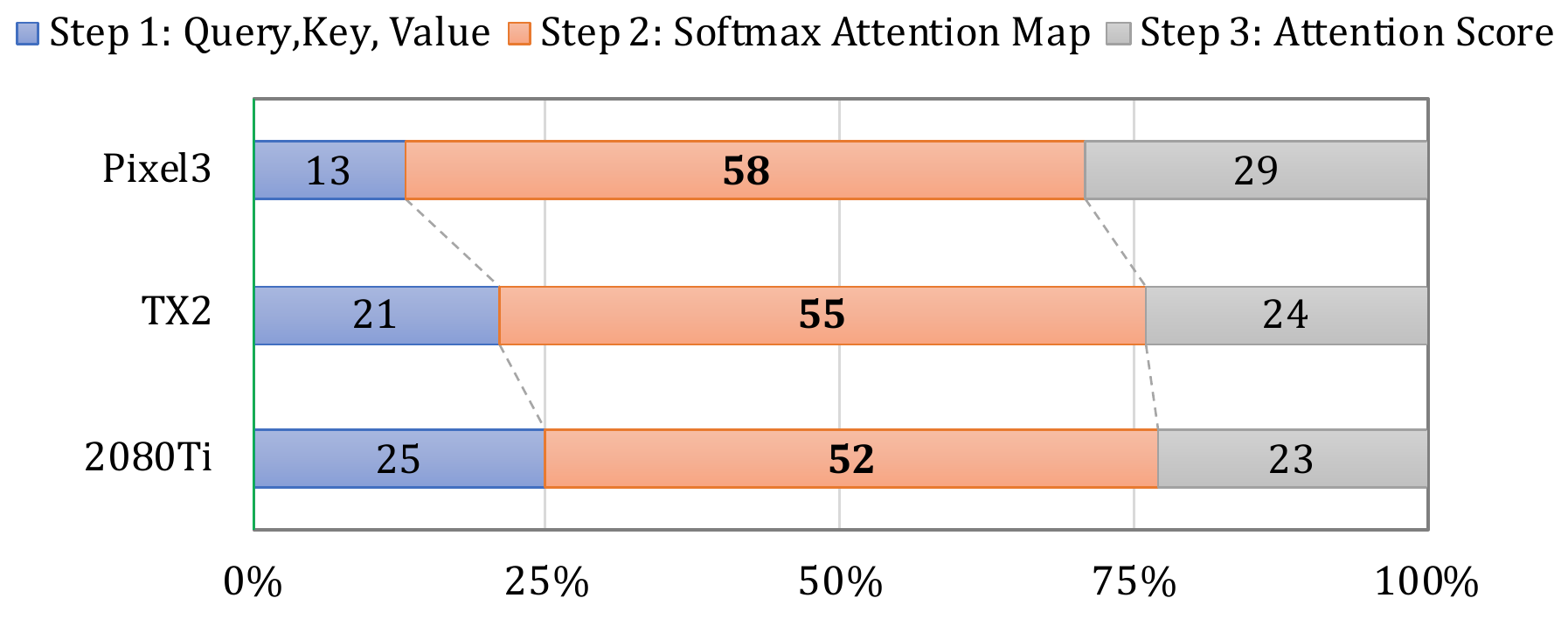}
    \vspace{-0.5em}
    \caption{Runtime breakdown of DeiT-Tiny MHA on various devices.}
    \label{fig:MHAprofiling}
    \vspace{-0.5em}
\end{figure}

To better understand the runtime breakdown for ViTs' MHA module, we profile DeiT-Tiny~\cite{pmlr-v139-touvron21a}, a popular ViT model, on various commercial devices, such as NVIDIA RTX 2080Ti~\cite{2080ti}, NVIDIA Edge GPU TX2~\cite{tx2}, and Google Pixel3 phone~\cite{pixel3}. 
In Fig. ~\ref{fig:MHAprofiling}, we observe that computing the softmax attention (Step 2) consistently dominates ($52\%-58\%$) the MHA runtime, especially when devices become less powerful and more resource-constrained. Hence, the major bottleneck for ViTs is the softmax attention, which limits their achievable efficiency and scalability, and prohibits extensive real-world ViT applications on resource-constrained devices.

To alleviate the above quadratic complexity, a simple approach is to reduce the number of patches or input resolution. However, this would result in larger patch sizes with an additional burden on hardware resources to compute the corresponding queries, keys, and values (\textit{Step 1 in Fig.~\ref{fig:MHAprofiling}}). Moreover, many real-world computer vision applications, such as medical imaging, autonomous driving, drone imagery and surveillance, etc, require high resolution inputs to discover finer-grained details in the images~\cite{https://doi.org/10.48550/arxiv.2205.14756}. 

A popular alternative approach in Transformers for Natural Language Processing (NLP) is to use sparse attention. Here, the attention matrix generated by the dot product between the dense queries and keys is made sparse by using a binary mask. Then, the sparse attention matrix with reduced entries goes through the softmax operation after which it is multiplied by a dense value matrix. Many works in algorithm~\cite{beltagy2020longformer, child2019generating, zaheer2020big, correia2019adaptively, cui2019fine, zhao2019explicit} and hardware~\cite{ham20203, wang2021spatten, sanger, qu2022dota,ham2021elsa} have been proposed to implement such sparse attentions for NLP-based Transformer models by efficiently tackling various static and dynamic sparse patterns.

 Orthogonal to the aforementioned techniques, we replace the vanilla softmax attention with a linear attention, and leverage the matrix associative property to linearize the cost of computing ViTs' attentions. Such linear attention has been proposed for NLP-based Transformer models~\cite{katharopoulos2020transformers,choromanski2021rethinking}. However, there is a missed opportunity in applying linear attentions to ViTs. Unlike sparsity-based accelerators, we seek to exploit the low-rank property of the proposed linear attention to design dedicated accelerator for improved latency and energy efficiency. Our main contributions are summarized below:
 \begin{enumerate}
   \item We propose an algorithm-accelerator codesign framework, dubbed \textsc{ViTALiTy}, that unifies low-rank and sparse approximation to boost the achievable accuracy-efficiency of ViTs using linear Taylor attentions. To the best of our knowledge, this is first of its kind work dedicated for ViTs by exploiting the low-rank properties in linear attentions.

    \item On the algorithm level, we propose a linear attention for reducing the computational and memory cost by decoupling the vanilla softmax attention into its corresponding ``weak'' and ``strong'' Taylor attention maps. Unlike the vanilla attentions, the linear attention in \textsc{ViTALiTy} generates a \textit{global context matrix} $\mathbf{G}$ by multiplying the keys with the values. Then, we unify the low-rank property of the linear attention with a sparse approximation of ``strong'' attention for training the ViT model. Here, the low-rank component of our \textsc{ViTALiTy} attention captures \textit{global information} with a linear complexity, while the sparse component boosts the accuracy of linear attention model by enhancing its \textit{local feature} extraction capacity.
    
    \item At the hardware level, we develop a dedicated accelerator to better leverage the algorithmic properties of \textsc{ViTALiTy}'s linear attention, where \textit{only a low-rank component} is executed during inference favoring hardware efficiency. Specifically, \textsc{ViTALiTy}'s accelerator features a chunk-based design integrating both a systolic array tailored for matrix multiplications and pre/post-processors customized for \textsc{ViTALiTy} attentions' pre/post-processing steps. Furthermore, we adopt an intra-layer pipeline design to leverage the intra-layer data dependency for enhancing the overall throughput, together with a down-forward accumulation dataflow for the systolic array to improve hardware efficiency. 
   
   \item We perform extensive experiments and ablation studies to demonstrate the effectiveness of \textsc{ViTALiTy} in terms of latency speedup ($3\times$), and energy efficiency ($3\times$) under comparable model accuracy with respect to the state-of-the-art solution.
\end{enumerate}

\section{Background and Motivation}

\subsection{Preliminaries of Vision Transformers }
\begin{figure}[t]
    \centering
    \includegraphics[width=\linewidth]{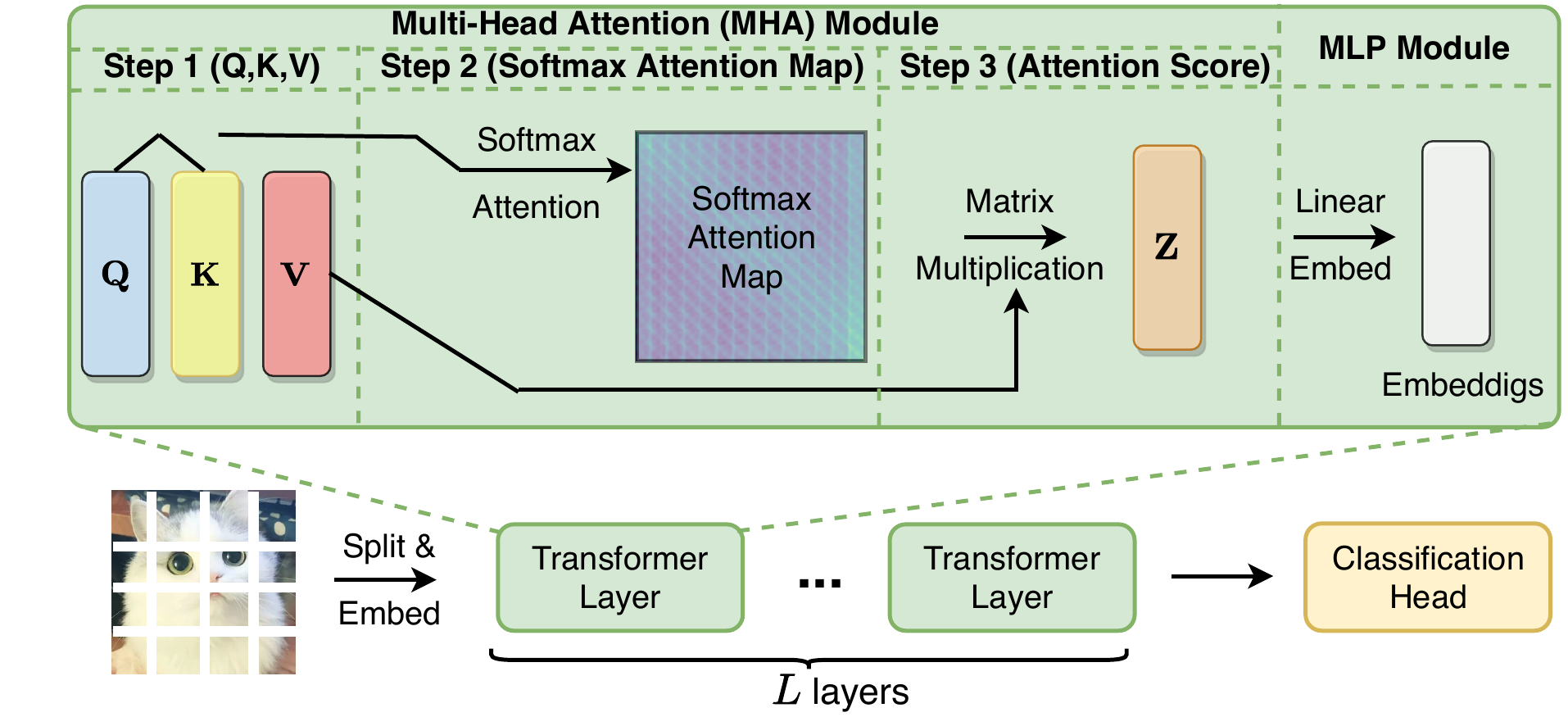}
    \caption{
   Each Transformer layer in a ViT model comprises a Multi-Head Attention (MHA) module and a Multi-Layer Perceptron (MLP) module.} 
    \label{fig:vit_arch}
    \vspace{-1em}
\end{figure}

\textbf{ViT Model Architecture.} 
Fig.~\ref{fig:vit_arch} illustrates the model architecture for ViTs.  Here, each input image is divided and arranged into a sequence of patches (or tokens), which are then fed into an $L$-layer Transformer encoder~\cite{vaswani2017attention}. Each Transformer layer comprises a multi-head attention (MHA) module and a multi-layer perceptron (MLP) module. 
As an example, the DeiT-Tiny model~\cite{pmlr-v139-touvron21a} consists of $L=12$ Transformer layers where the typical input image resolution is $224 \times 224$ with patch size $16 \times 16$; This results in a sequence of $n=196$ patches (tokens) with each token embedded as $64 \times 3$ with $h=3$ heads, and $d=64$ dimensions per head.

\vspace{0.5em}

\textbf{Attentions in ViTs.} The MHA module enables ViTs' impressive success in various tasks by enhancing the model's capacity to capture global information as compared to CNNs. Specifically, MHA receives $\mathbf{X} \in \mathbb{R}^{n \times d}$ as its input and outputs $\mathbf{Z} \in \mathbb{R}^{n \times d}$ as the attention score, which involves the following three computational steps as shown in Fig. ~\ref{fig:vit_arch}.

\vspace{0.3em}
\textbf{Step 1: Compute the query, key, value vectors} \\
\vspace{-0.5em}
$$\mathbf{Q} = \mathbf{X}\mathbf{W}^Q, \mathbf{K} = \mathbf{X}\mathbf{W}^K, \mathbf{V} = \mathbf{X}\mathbf{W}^V,$$
where $\mathbf{Q}$, $\mathbf{K}$, $\mathbf{V} \in \mathbb{R}^{n \times d}$ are the input embeddings for the next step. $\mathbf{W}^Q, \mathbf{W}^K,\mathbf{W}^V \in \mathbb{R}^{d \times d}$ are learned weights.


\textbf{Step 2: Compute the softmax attention map} 
$$\mathbf{S} = \texttt{softmax}\Big(\frac{\mathbf{Q} \mathbf{K}^T}{\sqrt{d}}\Big)$$

\textbf{Step 3: Compute the attention score}, $\mathbf{Z} = \mathbf{S} \mathbf{V}$

\noindent
Before moving to the next layer, the attention scores are sent to the MLP module,
$\mathbf{O} = \mathbf{Z}\mathbf{W}^O, \text{where, } \mathbf{W}^O \in \mathbb{R}^{d \times d}$.
\subsection{Related Work }
\textbf{Efficient Vision Transformers.}
Motivated by the breakthroughs of Transformers~\cite{vaswani2017attention, devlin2018bert, radford2018improving, liu2021Swin} in NLP, there has been a growing interest in developing Transformers for vision tasks.  ViT~\cite{vit} was the first to show that Transformers can completely replace convolutions by treating images as a sequence of patches of fixed length. Since then, ViT models and their variants have been successfully used for image recognition~\cite{vit, pmlr-v139-touvron21a}, object detection~\cite{carion2020end, zhu2020deformable}, and segmentation~\cite{ye2019cross}.
To capture fine-grained spatial details of different scales, multi-stage
hierarchical ViTs, such as Pyramid ViT~\cite{wang2021pyramid}, Swin Transformer~\cite{liu2021Swin}, Focal Transformer~\cite{yang2021focal}, and CrossViT~\cite{chen2021crossvit}, have been proposed, where the number of tokens is gradually reduced while the token feature dimension is progressively increased. Recent works in deploying efficient ViTs have sparked great interest. For example, by replacing local processing in convolutions with global processing via attention, MobileViT~\cite{mehta2021mobilevit} strives to combine the strengths of CNNs and ViTs;  LeViT~\cite{Graham_2021_ICCV} adopts a hybrid neural architecture. In parallel, compact ViT models via pruning~\cite{zhu2021visual} have been developed by encouraging dimension-wise sparsity; \cite{vit_quantization} proposes a post-training mixed-precision quantization scheme for reducing ViTs' memory storage and computational costs; and neural architecture search (NAS) has been adopted to discover better ViT models, e.g., NASViT~\cite{nasvit} is derived from Supernet-based one-shot NAS.

\textbf{Linear Attentions.} To alleviate the quadratic complexity associated with computing and storing attentions in Transformers, linear attention~\cite{katharopoulos2020transformers,choromanski2021rethinking,shen2021efficient} replaces the softmax operation with a generic similarity function defined as $\texttt{sim}(\mathbf{Q}, \mathbf{K}) = \phi(\mathbf{Q}) \phi(\mathbf{K})^T$, where $\phi()$ is a kernel or low-rank function. The above formulation exploits the matrix associative property to compute $\phi(\mathbf{Q})\big(\phi(\mathbf{K})^T\mathbf{V}\big)$, thereby reducing the quadratic computational complexity in vanilla Transformer attentions to a linear one. For NLP, Linear Transformer~\cite{katharopoulos2020transformers} defines a kernel function $\phi() = \texttt{elu}()+1$, while Performer~\cite{choromanski2021rethinking} uses positive orthogonal random features (PORF) as a low-rank function $\phi()$. Scatterbrain~\cite{chen2021scatterbrain} shows that combining both low-rank (via kernel feature map in Performer~\cite{choromanski2021rethinking}) linear attention, and sparse attention (via locality sensitive hashing in Reformer~\cite{Kitaev2020Reformer}) leads to efficient approximation with better performance than the individual components. For vision tasks, Efficient Attention~\cite{shen2021efficient} uses $\phi() = \texttt{softmax}()$ separately on queries and keys to approximate the vanilla softmax attentions. 

\textbf{Transformer Accelerators.}
There has been a surge in software-hardware co-designed accelerators dedicated to NLP Transformers which leverage dynamic sparsity patterns to tackle the quadratic complexity in storing and computing the attentions, e.g., $A^3$~\cite{ham20203}, SpAtten~\cite{wang2021spatten}, Sanger~\cite{lu2021sanger}, ELSA~\cite{ham2021elsa}, and DOTA~\cite{qu2022dota}. Specifically, $A^3$~\cite{ham20203} greedily searches for key vectors which are most relevant to the current query vector for approximating the vanilla attentions, but can suffer from low speedup and poor accuracy under high sparsity ratios; SpAtten~\cite{wang2021spatten} attempts to remove attention heads and tokens via structured pruning, which may still contain redundant attentions, leading to a low achievable sparsity; Sanger~\cite{lu2021sanger} dynamically sparsifies the vanilla attentions based on a quantized prediction of the attentions followed by rearranging the sparse mask via a ``pack and split” strategy into hardware-friendly structured blocks; ELSA~\cite{ham2021elsa} adopts binary hashing maps to estimate the angles between the queries and keys for approximating attentions, for enabling lightweight similarity computation together with a specialized accelerator, which can suffer from a degraded accuracy; and DOTA~\cite{qu2022dota} adopts both low precision and low-rank linear transformation to predict the sparse attention masks by jointly optimizing a lightweight detector together with the Transformer to accurately detect and omit weak connections during runtime.
\subsection{Gaps and Opportunities}
In contrast to existing dynamic sparsity-based accelerators dedicated to NLP Transformers, there still exists a missing gap for ViT accelerators. Unlike NLP Transformers and their corresponding accelerators which deal with input sequences of varying lengths during runtime, ViTs only needs to handle a fixed-length input sequence (tokens) of image patches. Hence, there is an opportunity to leverage this unique property to boost the efficiency of ViT acceleration. 

Overall, existing works on efficient ViT models focus on the model architectures, while existing Transformer accelerators target sparse attentions of NLP Transformers. Hence, a systematic counterpart approach is still missing for accelerating ViT models. Moreover, there is still a lack of works exploiting low-rank properties of attentions for designing ViT accelerators with boosted hardware efficiency. Finally, there is a great potential of training ViT models with combination of low-rank linear attentions and sparse attentions for restoring the accuracy of linear attention ViTs. To close the above gaps, we propose a first-of-its-kind algorithm-hardware co-designed framework dedicated to ViT inference via linear Taylor attentions. Unlike the Scatterbrain~\cite{chen2021scatterbrain} algorithm, we decouple vanilla softmax attentions as a combination of low-rank and sparse approximations during training to boost the  accuracy of linear attention ViTs, and drop the sparse attention while retaining the linear attention during inference to avoid any run-time overhead associated with dynamic sparse attentions.
\section{Proposed \textsc{ViTALiTy} Algorithm }
\subsection{Distribution of ViT Attentions}
\vspace{-0.3em}
Here we first analyze the distribution of ViT attention values under row-wise mean-centering, and then present a motivating observation for our proposed \textsc{ViTALiTy} algorithm. To do so, we begin by reviewing a key property of softmax functions with mean-centered input sequences. 
\begin{property}[Mean-Centering]\label{prop1}
Given a sequence of $n$ data samples denoted as $\mathbf{x}=\{x_i,\; i= 1,\ldots,n\}$ with a scalar mean $\overline{x}$, it can be analytically shown that subtracting a scalar value does not change the softmax output.
\[ \texttt{softmax}(\mathbf{x}-\overline{x})_{i} = \frac{\texttt{exp}({x_i-\overline{x})}}{\sum_{i=1}^{n}\texttt{exp}({x_i-\overline{x})}} = \texttt{softmax}(\mathbf{x})_i\]
\end{property}
We seek to apply the above property to the scaled dot product attention (similarity) matrix, $\frac{\mathbf{QK^T}}{\sqrt{d}}$, as the input to the softmax function. 
However, computing row-wise mean-centering of the dot product attention matrix for each head is computationally expensive, as it relies on first computing, and storing the attention which is \textit{quadratic} in $n$. 

\textbf{Proposed Efficient Mean-Centering Attentions.}  
To alleviate the above challenge for all rows $i\in[n]$, we propose to efficiently compute the mean-centering rows of the attention in \textit{linear} time by directly modifying the key matrix, $\mathbf{K}$.  
\begin{flalign*}
& \frac{\mathbf{QK^T}}{\sqrt{d}} - \; \texttt{mean}\bigg( \frac{\mathbf{QK^T}}{\sqrt{d}} \bigg) =
\frac{\mathbf{Q}}{\sqrt{d}} \big(\mathbf{K}-\mathbf{1}_n\overline{\mathbf{K}}\big)^T =
\frac{\mathbf{Q}\Hat{\mathbf{K}}^T}{\sqrt{d}}, \\
& \text{where, } \overline{\mathbf{K}} =
\frac{1}{n}\sum_{j=1}^{n}\mathbf{k}_j = \frac{1}{n} \big(\mathbf{1}_{n}^{T}\mathbf{K}\big)
\end{flalign*}
$\Hat{\mathbf{K}}$ is the mean-centered key matrix that allows us to avoid computing, and storing the more expensive $\mathbf{QK^T}$ attention for each head prior to performing mean-centering.
\vspace{0.2em}
The key benefit of mean-centering is to regularize the softmax inputs to be centered around zero so that their distribution is geared towards a normal distribution curve for a majority number of inputs (Central Limit Theorem), thereby ensuring that more samples fall within the interval $[-1, 1)$.
By leveraging Property~\ref{prop1} for each row of an attention matrix, we get
  \[ \texttt{softmax}\bigg(\frac{\mathbf{Q}\Hat{\mathbf{K}}^T}{\sqrt{d}}\bigg) = \texttt{softmax}\bigg(\frac{\mathbf{QK^T}}{\sqrt{d}}\bigg),  \]
which transforms the similarity between the queries and modified keys, $(\mathbf{q}_i\Hat{\mathbf{k}}_j^T)$, to be centered around zero-mean without changing the softmax outputs.
\begin{figure}[t]
    \centering
    \includegraphics[width=\linewidth]{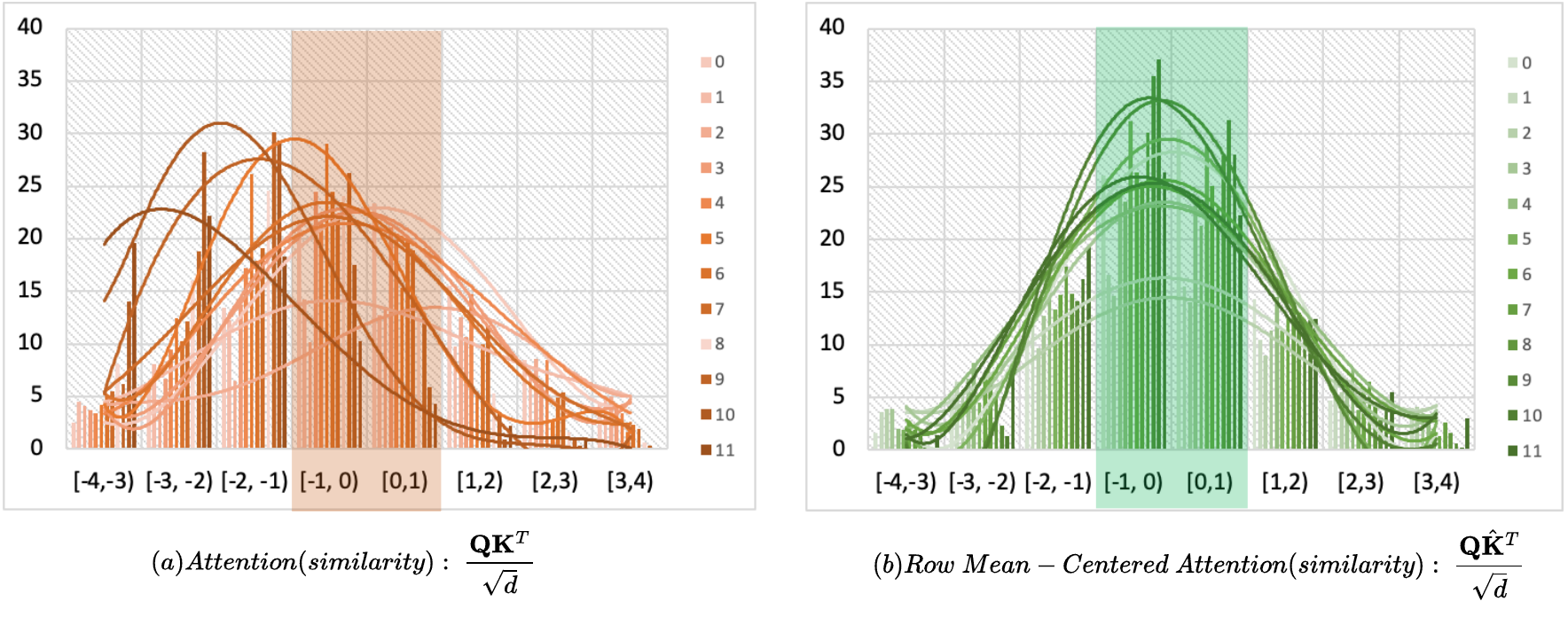}
    \caption{Distribution of attentions (inputs to softmax) across various layers 0-11: (a) the vanilla attention distribution shifts left, and (b) row-wise mean-centering of attentions centers the distribution towards [-1,1). Here we use the DeiT-Tiny~\cite{pmlr-v139-touvron21a} model on the ImageNet dataset as an example.}
    \label{fig:distribution}
\end{figure}
\vspace{0.5em}

\textbf{Motivating Observation.} Fig. \ref{fig:distribution} visualizes the distribution of attention (similarity) matrices of various layers in the DeiT-Tiny ViT model with multiple heads on the ImageNet dataset, \textit{before} and \textit{after} mean-centering. We observe that up to 
$67\%$ entries of the mean-centered attention values lie within the interval $[-1, 1)$  compared 
with that of $46\%$ in the vanilla ones, i.e., 21\% increase after mean-centering. In other words, efficiently row-wise mean-centering the attentions centers majority of the similarity between the queries and modified keys around zero, thereby revealing denser weak connections. Furthermore, the similarity values lying outside the interval $[-1,1)$ are sparser in quantity but represent stronger connections between queries and keys. This motivates that the mean-centered softmax attention can be decoupled into a sum of ``weak'' and ``strong'' attentions which is realized by the proposed Taylor attention.
\subsection{Constructing Taylor Attentions for ViTs}
\label{sec:TA}
We define Taylor attention based on Taylor series expansion of the exponential function $\texttt{exp}()$ used in row-wise mean-centered softmax attention. We recall that up to $67\%$ of the similarity matrix lies within the interval $[-1,1)$ representing denser weak connections. Since Taylor series approximation for values close to zero can be represented with first-order Taylor expansion, we get $\texttt{exp}(\mathbf{q}_i \hat{\mathbf{k}}_j^T) \approx 1 + \mathbf{q}_i \hat{\mathbf{k}}_j^T$. 
However, a key challenge here is to accurately locate the weak (query, key) connections within the attention matrix for a given input. In concurrent work for NLP-based Transformer accelerators ~\cite{wang2021spatten, qu2022dota,ham2021elsa, lu2021sanger}, these weak connections are dynamically computed or predicted during runtime which are pruned resulting in irregular sparse attention patterns and complex designs for the corresponding accelerator. In contrast, for ViTs our novelty lies in decoupling the softmax attentions as a combination of ``weak'' and ``strong'' Taylor attention maps without any overheads of identifying the weak connections and generating dynamic sparse attention patterns during runtime.

The ``weak'' Taylor attention map is computed by the first-order ($m=1$) Taylor approximation of the original softmax attention matrix capturing the weak connections between $\mathbf{Q}$ and $\hat{\mathbf{K}}$. Let, $\hat{\mathbf{k}}_{sum} = \sum_{j=1}^n \hat{\mathbf{k}}_j = \mathbf{1}_{n}^T\Hat{\mathbf{K}}$.
\begin{flalign*}
& \scriptsize{\texttt{Taylor}_{softmax}\bigg(\frac{\mathbf{Q}\Hat{\mathbf{K}}^T}{\sqrt{d}}\bigg)\Bigg|_{m=1} \hspace{-1em}= \Bigg[\texttt{Taylor}_{softmax}\bigg(\frac{\mathbf{q}_i\Hat{\mathbf{K}}^T}{\sqrt{d}}\bigg)\Bigg]_{i \in [n]}} \\
& = \texttt{diag}^{-1} \big( n\sqrt{d}\mathbf{1}_n + \mathbf{Q}\hat{\mathbf{k}}_{sum}^T\big)\times \Big(\sqrt{d}\;\mathbf{1}_n\mathbf{1}_n^T + \mathbf{Q}\Hat{\mathbf{K}}^T \Big)
\end{flalign*}

However, directly replacing a softmax attention with the first-order Taylor attention in a pre-trained ViT models leads to poor accuracy, which we further validate in Fig.~\ref{fig:baseline} (\textsc{LowRank}). This degradation in accuracy is explained by incorrectly assuming that all (query, key) connections are weak with their similarity measure falling within the interval $[-1, 1)$, for various layers and heads across different inputs in a ViT model. Hence, it is imperative to include the corresponding higher-order ($m>1$) terms from the Taylor series expansion of $\texttt{exp}\big(\frac{\mathbf{Q}\hat{\mathbf{K}}^T}{\sqrt{d}}\big)$ to compensate for the (query, key) pairs with similarity outside the interval $[-1,1)$ representing ``strong'' connections by their magnitude. Hence, we decouple the vanilla softmax attention as follows:
\begin{flalign*}
& \texttt{softmax}\bigg(\frac{\mathbf{Q}\mathbf{K}^T}{\sqrt{d}}\bigg) = \texttt{softmax}\bigg(\frac{\mathbf{Q}\hat{\mathbf{K}}^T}{\sqrt{d}}\bigg) \\
& \small{ \approx \underbrace{\texttt{Taylor}_{softmax}\bigg(\frac{\mathbf{Q}\Hat{\mathbf{K}}^T}{\sqrt{d}}\bigg)\Bigg|_{m=1}}_{\text{"weak attention"}} \hspace{-1 em} + \underbrace{\texttt{Taylor}_{softmax}\bigg(\frac{\mathbf{Q}\Hat{\mathbf{K}}^T}{\sqrt{d}}\bigg)\Bigg|_{m>1}}_{\text{"strong attention"}}}
\end{flalign*} 

\subsection{Taylor Attention is a Double-edged Sword} Representing the vanilla softmax attention as a combination of ``weak'' and ``strong'' Taylor attention maps has both advantages and disadvantages which we discuss below.

\textbf{Advantages.} \ding{172} The ``weak'' Taylor attention represents a \textbf{linear attention} which enables exploiting the associative property of matrix multiplication for enabling a linear computational and memory complexity with the number of patches/tokens. Putting aside the normalization, the ``weak'' Taylor attention switches the order of the softmax attention from $\big(\mathbf{Q}\Hat{\mathbf{K}}^T\big)\mathbf{V}$  to more-efficient $\mathbf{Q}\big(\Hat{\mathbf{K}}^T\mathbf{V}\big)$. Though the final attention output dimension $n \times d$ remains the same, the computational time complexity reduces from quadratic $\mathbfcal{O}(n^2d)$ to linear $\mathbfcal{O}(nd^2)$. Consequently, there is \textit{no need} to compute and store the $\mathbfcal{O}(n^2)$ attention matrix $\big(\mathbf{Q}\Hat{\mathbf{K}}^T\big)$ for each query, rather a \textit{global context matrix}, $\mathbf{G} = \big(\Hat{\mathbf{K}}^T\mathbf{V}\big)$ is computed and stored with a memory cost of $\mathbfcal{O}(d^2)$, which is independent of the number of patches. Therefore, a linear Taylor attention can lead to both 1) a lower complexity and 2) a higher hardware utilization thanks to its resulting dense linear attention matrix.

\ding{173} Linear attention can enhance our understanding of softmax attentions via \textbf{low-rank} Taylor attention maps capturing various global contextual information. Here the \textit{global context matrix} $\mathbf{G}$ can be interpreted as a collection of distinct semantic aspects of the input aggregated by their corresponding values, i.e., different rows $\mathbf{g}_i$ correspond to a summary of distinct global context vectors that capture the foreground, core object, periphery, etc. This leads to enhanced understanding that the low-rank matrix, $\mathbf{QG}$ in the proposed (un-normalized) Taylor attention, $\sqrt{d}\big(\mathbf{1}_{n}\mathbf{v}_{sum}\big) + \mathbf{Q} \mathbf{G}$, is a linear combination, for which the weights are a set of coefficients in each query, of a set of global template attention maps, each having a semantically significant focus. In other words, pixels belonging to a particular semantic group might contribute a larger weighted coefficient to its corresponding global context vector in the linear attention, resulting in much refined score~\cite{shen2021efficient}. Moreover, $\sqrt{d}\big(\mathbf{1}_{n}\mathbf{v}_{sum}\big)$ is a rank-$1$ matrix and independent of the query and may represent the background attention. 
\vspace{0.5em}

\textbf{Disadvantages.} A key challenge in Taylor attention is computing the ``strong'' attention which represents the higher-order terms $m>1$ of Taylor expansion. Firstly, computing the optimal Taylor order for given input is non-trivial. Secondly, generating the higher-order Taylor attention maps requires multiplying the complete query and key matrices together for various orders which has quadratic complexity thereby eclipsing the advantages of linear attention. Therefore, we seek to estimate the ``strong'' attention based on sparse approximation.

\begin{figure*}[h]
    \centering
    \includegraphics[width=0.9\linewidth]{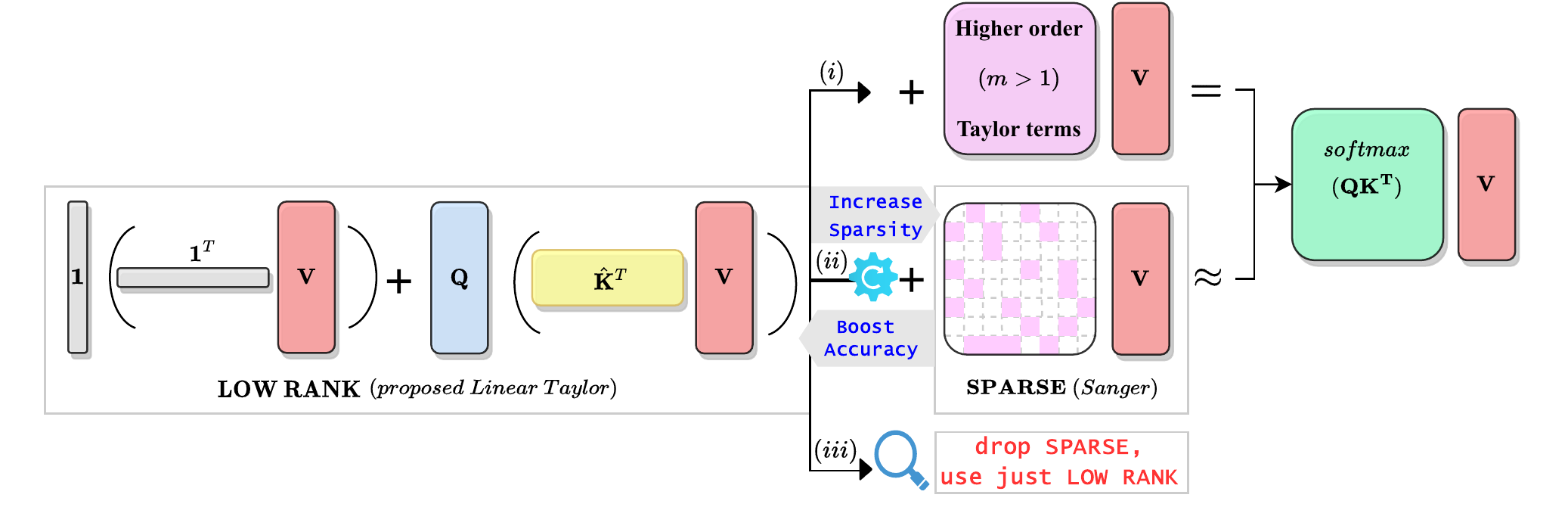}
   \caption{ \textsc{ViTALiTY}  workflow comprising the proposed (Low-Rank) Linear Taylor attention (order, $m=1$): (i) Higher-order Taylor terms ($m>1$) when added results in vanilla softmax attention score, (ii) Training phase (unifying low-rank and sparse approximation) where higher-order Taylor terms are approximated as Sparse attention (computed using \textsc{Sanger}~\cite{sanger}), and (iii) Inference phase that uses only the (Low-Rank) Linear Taylor attention. }
    \label{fig:vitality-workflow}
    \vspace{-1em}
\end{figure*}

\subsection{Proposed: Unifying Low-rank and Sparse Attentions}
As discussed earlier, combining the ``strong'' Taylor attention helps boost the achievable accuracy of ViTs against using only the ``weak'' Taylor attention. While the ``weak'' Taylor attention being a linear attention captures only the \textit{global context} without explicitly computing the attention matrix, it lacks the non-linear attention score normalization. Consequently, this weakens its ability to extract local features from high attention scores produced by local patterns in the input image resulting in degraded accuracy
~\cite{germain2022visual, Chen_2021_ICCV, tang2022quadtree, https://doi.org/10.48550/arxiv.2205.14756}. 
To enhance the local feature extraction capacity and boost the accuracy of linear attention ViTs, we seek to unify the above ``weak'' Taylor attention (dense low-rank) with the ``strong'' attention (sparse approximation) that captures the non-linear relationship between (queries, keys). 
In contrast to combining different works on low-rank and sparse components in Scatterbrain~\cite{chen2021scatterbrain}, \textsc{ViTALiTy} is unique as it decouples the vanilla softmax attention into the corresponding low-rank component given by the proposed linear Taylor attention, $\sqrt{d}\big(\mathbf{1}_{n}\mathbf{v}_{sum}\big) + \mathbf{Q} \mathbf{G}$, depicting global context, and the sparse component based on approximating the higher-order terms capturing the local features with ``strong'' connections as illustrated in Fig~\ref{fig:vitality-workflow}. We use the sparsity prediction mask generated by the quantized query and key following Sanger~\cite{sanger} for fine-tuning our models. Our key insights which we empirically validate in Section~\ref{sec:ablation} are:
\begin{enumerate}
    \item {We show that the widely used softmax attention could represent both low-rank and sparse components. \textbf{During training}, the low-rank component, i.e., linear Taylor attention allows the sparse attention to exhibit more sparsity with higher thresholds while the sparse component helps boost the accuracy of otherwise low-rank Taylor attention model.} 
    \item {\textbf{During inference}, we discover that using only the low-rank component, i.e., linear Taylor attention is sufficient as it exhibits similar accuracy to that of combined low-rank and sparse attention. This reveals that the sparse component acts as a regularizer to help boost the accuracy of linear Taylor attention during training, and hence can be dropped during inference to avoid overhead of dynamic sparse attentions.} 
\end{enumerate}
\begin{algorithm}[t]
\caption{\textsc{ViTALiTy} with Taylor Attention\label{alg:vitality}}
\KwInput{$\mathbf{Q}$, $\mathbf{K}$, $\mathbf{V}$ are queries, keys, values}
\KwOutput{$\mathbf{Z}$ is the Taylor attention score}
\Comment*[h]{Step 1: Mean-centering keys}


\quad $\overline{\mathbf{K}} \gets \frac{1}{n} \big(\mathbf{1}_{n}^{T}\mathbf{K}\big)$, \textcolor{OliveGreen}{\texttt{cost:}$\mathbfcal{O}(nd)$}

\quad $\Hat{\mathbf{K}} \gets \mathbf{K} - \mathbf{1}_{n} \overline{\mathbf{K}}$, \textcolor{OliveGreen}{\texttt{cost:}$\mathbfcal{O}(nd)$}

\Comment*[h]{Step 2: Global context matrix}

\quad $\mathbf{G} \gets \Hat{\mathbf{K}}^T \mathbf{V}$, \textcolor{OliveGreen}{\texttt{cost:}$\mathbfcal{O}(nd^2)$}

\Comment*[h]{Step 3: Column sum of keys, values}

\quad $\hat{\mathbf{k}}_{sum} \gets \mathbf{1}_{n}^T\Hat{\mathbf{K}}$, \textcolor{OliveGreen}{\texttt{cost:}$\mathbfcal{O}(nd)$}

\quad $\mathbf{v}_{sum} \gets \mathbf{1}_{n}^T \mathbf{V}$, \textcolor{OliveGreen}{\texttt{cost:}$\mathbfcal{O}(nd)$}

\Comment*[h]{Step 4: Compute Taylor denominator}

\quad $\mathbf{t}_{D} \gets \big(n\sqrt{d}\big)\mathbf{1}_{n} + \mathbf{Q} \hat{\mathbf{k}}_{sum}^T$, \textcolor{OliveGreen}{\texttt{cost:}$\mathbfcal{O}(nd)$}

\Comment*[h]{Step 5: Compute Taylor numerator}

\quad $\mathbf{T}_{N} \gets \sqrt{d}\big(\mathbf{1}_{n}\mathbf{v}_{sum}\big) + \mathbf{Q} \mathbf{G}$, \textcolor{OliveGreen}{\texttt{cost:}$\mathbfcal{O}(nd^2)$}

\Comment*[h]{Step 6: Taylor attention score}

\quad $\mathbf{Z} \gets \texttt{diag}^{-1} \big(\mathbf{t}_D\big)\mathbf{T}_{N}$, \textcolor{OliveGreen}{\texttt{cost:}$\mathbfcal{O}(nd)$}
\end{algorithm}

\begin{figure} [t]
    \centering
    \includegraphics[width=\linewidth]{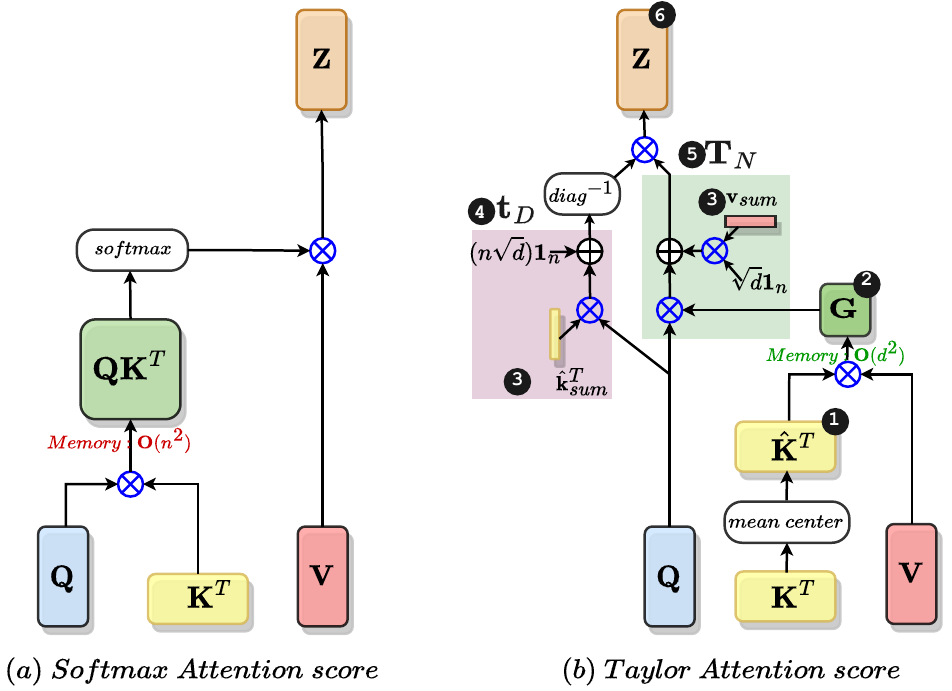}
   \caption{Computational steps (a) vanilla Softmax Attention and (b) our Taylor attention (see Algorithm \ref{alg:vitality}), where the \textit{global context matrix} $\mathbf{G}$ provides \textit{linear} computation and memory benefits over the vanilla \textit{quadratic} $\mathbf{Q}\mathbf{K}^T$.}
    \label{fig:workflowcomparison}
    \vspace{-1em}
\end{figure}

Algorithm~\ref{alg:vitality} presents the inference computational cost of \textsc{ViTALiTy} having a linear dependency on $n$, and Fig.~\ref{fig:workflowcomparison} compares the computational workflow of the proposed Taylor attention score with that of the vanilla softmax attention score.

\subfile{arch}
\subfile{results}

\section{Conclusions}
We present a new algorithm and hardware co-designed ViT framework dubbed \textsc{ViTALiTy} that unifies the low-rank linear Taylor attention with a sparse approximated attention \textit{during training} for boosting the achievable accuracy and then drops the sparse component \textit{during inference} for improving the acceleration efficiency without hurting the accuracy. 
Orthogonal to existing sparsity-based accelerators dedicated to NLP, our \textsc{ViTALiTy} accelerator is developed to better exploit the resulting algorithmic properties of \textsc{ViTALiTy}'s linear attention for boosting hardware efficiency during inference. \textsc{ViTALiTy} achieves up to $\textbf{3}\times$ end-to-end latency speedup with $\textbf{3}\times$ energy efficiency over Sanger~\cite{sanger} under a comparable accuracy.

\section*{Acknowledgements}
The work is supported by the NSF CAREER award (Award number: 2048183), and Meta Faculty Research Award on AI System Hardware/Software Codesign.


\newpage
\bibliographystyle{IEEEtranS}
\bibliography{references}

\end{document}

%% file: arch.tex
\section{Proposed \textsc{ViTALiTy} Accelerator}
Here, we first analyze the computational complexity of the proposed Taylor attention
, then discuss potential acceleration opportunities for leveraging our proposed attention, 
and finally present our developed \textsc{ViTALiTy} accelerator.
\subsection{Complexity Analysis and Potential Opportunities}
\textbf{Theoretical Complexity Analysis.}
Here we discuss the theoretical complexity in terms of the ratios of operation numbers, $R$, between the vanilla softmax attention and the proposed Taylor attention. Recall, $n$ and $d$ denote the number of tokens and the feature dimensions. In Eqs. (\ref{eq:mul}), (\ref{eq:add}), and (\ref{eq:div}), we compute the theoretical ratio of number, $R$, for multiplications (Mul.), additions (Add.), and divisions (Div.), respectively, and we conclude that:
1) In the vanilla softmax attention, the numbers of both multiplications (Mul.) (i.e., $\mathbf{QK}^T$ and $\mathbf{SV}$) 
and divisions (Div.) are nearly $(\mathbf{n} / \mathbf{d})\times$ more than those in our proposed Taylor attention (i.e., $\mathbf{G}=\Hat{\mathbf{K}}^T \mathbf{V}$, $\mathbf{Q} \hat{\mathbf{k}}_{sum}^T$, and $\mathbf{Q}\mathbf{G}$ for Mul., and Steps 1 and 6 in Algorithm \ref{alg:vitality} for Div.);
2) While the number of addition (Add.) reduction in our Taylor attention compared to the vanilla attention is less than $(\mathbf{n}/\mathbf{d})\times$ owing to extra pre-processing steps (see Steps 1 and 3 in Algorithm \ref{alg:vitality}); 
3) Furthermore, in contrast to the vanilla attention, our Taylor attention does not have the computationally expensive exponentiation (Exp.).
\begin{equation}
\vspace{-0.5em}
	R_{Mul.} = \frac{2\times n \times n\times d}{(2\times n\times d\times d) +( n\times d)}
	= \frac{2n}{2d + 1}
	\approx \frac{n}{d}, \label{eq:mul}
\end{equation}
\begin{equation}
	R_{Add.} = \frac{2\times n^{2}\times d + n^{2}}{(2\times n\times d^{2}) + (7nd)}
	= \frac{(2d + 1)n}{(2d + 7)d}
	< \frac{n}{d}, \label{eq:add}
\end{equation}
\vspace{-0.2cm}
\begin{equation}
	R_{Div.} = \frac{n\times n}{nd + d}
	= \frac{n^{2}}{(n+1)d}
	\approx \frac{n}{d}. \label{eq:div}
\end{equation}

\begin{table}[t]
\centering
\caption{Comparing operation numbers (M) between \textsc{ViTALiTy} Taylor attention and vanilla Softmax attention on ViT models.}
\setlength{\tabcolsep}{0.3em}
\resizebox{\linewidth}{!}{
\begin{tabular}{|c||c|c|c||cc|cc|c|cc|}
\hline
\multicolumn{1}{|l||}{} & \multicolumn{3}{|c||}{\textsc{\textbf{ViTALiTy}}} & \multicolumn{7}{c|}{\textsc{\textbf{Baseline}} (vanilla softmax)}                                                    \\ \hline
\textbf{MODELS}                & {\textbf{Mul.}}       & {\textbf{Add.}}       & {\textbf{Div.}}       & \multicolumn{2}{c|}{\textbf{Mul.}} & \multicolumn{2}{c|}{\textbf{Add.}} & \textbf{Exp.} & \multicolumn{2}{c|}{\textbf{Div.}} \\
\hline
DeiT-Tiny             & $58.3$         & $61.0$       & $0.5$        & $178.8$     & $(3.1\times)$    & $180.2$    & $(3.0\times)$   & $1.4$  & $1.4$     & $(3.1\times)$    \\ \hline
MobileViT-xs        & $4.8$          & $5.3$        & $0.1$        & $28.4$      & $(5.9\times)$     & $29.0$     & $(5.5\times)$  & $0.6$  & $0.6$     & $(5.1\times)$     \\ \hline
LeViT-128             & $3.4$          & $4.0$        & $0.1$        & $36.4$      & $(10.7\times)$    & $37.5$     & $(9.4\times)$  & $1.1$  & $1.1$     & $(10.6\times)$    \\ \hline
\end{tabular}} \label{tab:flops}
\vspace{-0.4cm}
\end{table}    

\noindent Table \ref{tab:flops} empirically validates above conclusions, where we observe that for any given model, the ratios of numbers for each of the multiplications, additions, and division are nearly similar. Moreover, for various models, i.e.,  DeiT-Tiny \cite{pmlr-v139-touvron21a}, MobileViT-xs \cite{mehta2021mobilevit}, and LeViT-128 \cite{Graham_2021_ICCV}, we find the ratio of operation numbers to be about $\textbf{3}\times$, $\textbf{5}\times$, and $\textbf{10}\times$, respectively.
\vspace{-0.5em}

\textbf{Profiling Results Analysis.}
Despite the theoretical reduction of operation numbers in \textsc{ViTALiTy}'s Taylor attention as discussed above, existing general computing platforms or accelerators dedicated to the vanilla ViT's attention cannot fully take advantage of our Taylor attention's potential benefits to boost its achievable hardware efficiency.
For example, Table \ref{tab:profiling} compares the profiling latency of both the proposed and vanilla attentions in several representative ViTs, including DeiT-Tiny \cite{pmlr-v139-touvron21a}, MobileViT-xs \cite{mehta2021mobilevit}, and LeViT-128 \cite{Graham_2021_ICCV} on a typical edge GPU (i.e., NVIDIA Tegra X2), and we can see that:
1) The softmax operation, which accounts for about $40\%$ of the overall latency, is arguably the dominant operator in the vanilla attention;
2) Although \textsc{ViTALiTy} adopts the Taylor attention to remove the hardware inefficient softmax operation and further reduce the number of multiplications/additions/divisions, its potential hardware efficiency does not reflect in the profiling latency on the GPU, motivating dedicated accelerators for our \textsc{ViTALiTy} algorithm; 
3) As each step in Algorithm \ref{alg:vitality} is processed sequentially on the GPU without exploring potential pipeline opportunities, even the computationally light pre/post-processing steps contribute to a nontrivial amount of the overall latency.
\begin{table}[t]
\centering
\caption{The latency profiling results of \textsc{ViTALiTy}'s Taylor attention and the vanilla attention on the edge GPU (NVIDIA Tegra X2).}
\setlength{\tabcolsep}{0.3em}
\resizebox{\linewidth}{!}{
\begin{tabular}{|l|cc|cc|cc|}
\hline
                                        & \multicolumn{2}{c|}{\textbf{DeiT-Tiny} \cite{pmlr-v139-touvron21a}}                                       & \multicolumn{2}{c|}{\textbf{MobileViT-xs} \cite{mehta2021mobilevit}} & \multicolumn{2}{c|}{\textbf{LeViT-128} \cite{Graham_2021_ICCV}}                                              \\ \hline
\textsc{\textbf{ViTALiTy}}(Taylor attention)                     & \textbf{Latency}(ms)    & \textbf{Ratio}                                               & \textbf{Latency}(ms)             & \textbf{Ratio}      & \textbf{Latency}(ms)       & \textbf{Ratio}                                               \\ \hline
1: $\hat{\mathbf{K}}$                  & $1.40$           & $10\%$ & $0.41$                & $15\%$       & $0.65$          & $15\%$ \\
2:  $\mathbf{G}$                          & $3.51$           &  $25\%$ & $0.62$                & $22\%$       & $1.05$          & $24\%$ \\
3:  $\mathbf{\hat{k}}_{sum}$, $\mathbf{v}_{sum}$ & $1.40$           & 10\%                                                & $0.41$                & $15\%$       & $0.65$          & $15\%$                                                \\
4:  $\mathbf{t}_{D}$                    & $2.24$           & $16\%$                                                & $0.41$                & $15\%$       & $0.65$          & $15\%$                                                \\
5:  $\mathbf{T}_{N}$                    & $3.93$           & $28\%$                                                & $0.64$                & $23\%$       & $1.11$          & $25\%$                                                \\
6:  $\mathbf{Z}$                          & $1.54$           & $11\%$                                                & $0.27$                & $10\%$       & $0.32$          & $7\%$                                                 \\ 
OVERALL                               & $\textbf{14.03}$ & $100\%$                                               & $\textbf{2.76}$       & 100\%      & $\textbf{4.43}$ & $100\%$                                               \\ \hline
\textsc{\textbf{Baseline}} (vanilla softmax)                      & \textbf{Latency}(ms)    & \textbf{Ratio}                                               & \textbf{Latency}(ms)             & \textbf{Ratio}      & \textbf{Latency}(ms)       & \textbf{Ratio}                                               \\ \hline
1. $\mathbf{QK}^{T}$                           & $3.61$           & $31\%$                                                & $0.54$                & $30\%$       & $0.80$          & $29\%$                                                \\
2. $\mathbf{S}=\texttt{Softmax}(\mathbf{QK}^{T})$                & $4.43$           & $38\%$                                                & $0.72$                & $40\%$       & $1.16$          & $42\%$                                                \\
3. $\mathbf{SV}$                                 & $3.61$           & $31\%$                                                & $0.54$                & $30\%$       & $0.80$          & $29\%$                                                \\
OVERALL                              & $\textbf{11.65}$ & 100\%                                               & $\textbf{1.79}$       & $100\%$      & $\textbf{2.76}$ & $100\%$                                               \\ \hline
\end{tabular}} \label{tab:profiling}
\vspace{-1em}
\end{table}
Based on the theoretical and profiling analysis above, we next discuss potential opportunities of leveraging our \textsc{ViTALiTy} algorithm's properties for designing dedicated accelerators with boosted hardware efficiency, which motivate our accelerator design in Section \ref{sec:micro-arch} and Section \ref{sec:execution_steps}. 

\begin{figure*} [!t]
    \centering
    \includegraphics[width=0.9\linewidth]{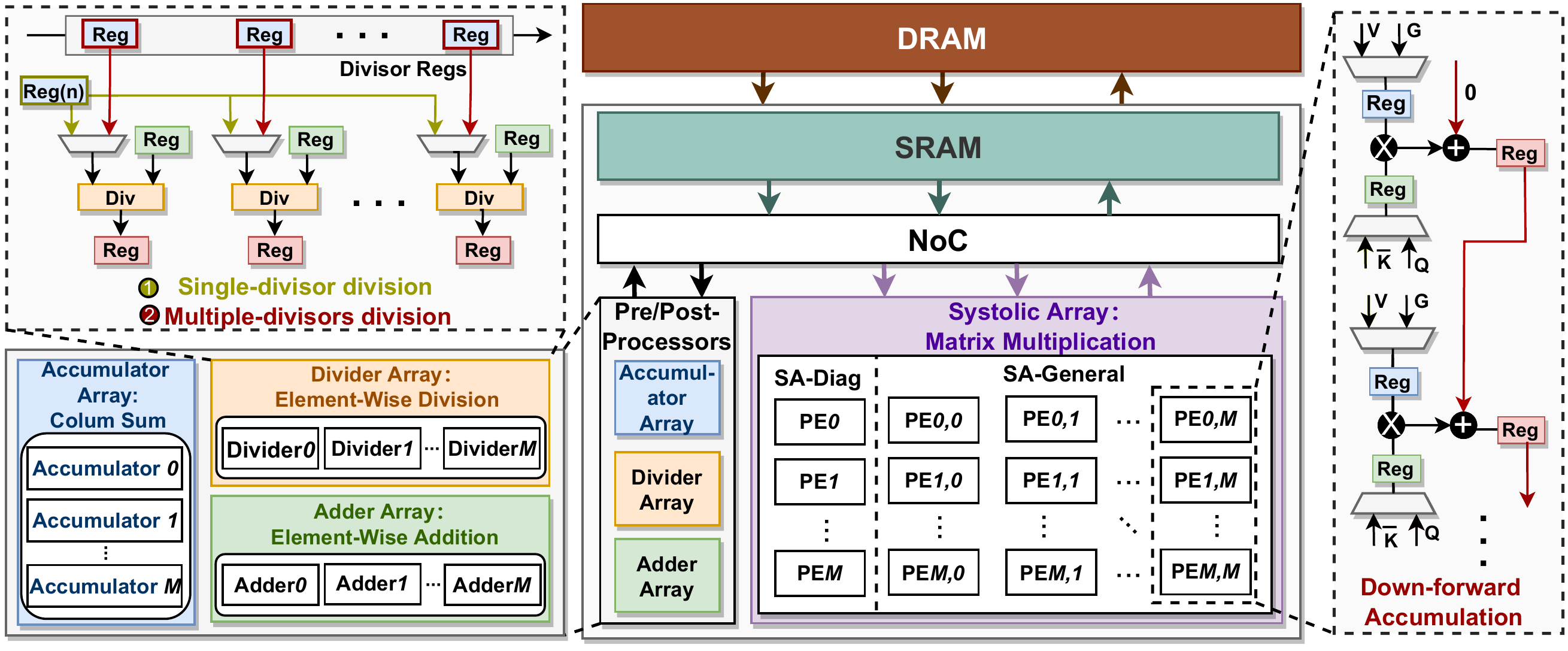}
    \vspace{-0.6em}
    \caption{An illustration of our \textsc{ViTALiTy} accelerator, which adopts four memory hierarchies (i.e., DRAM, SRAM, NoC, and Regs) to enhance data locality and multiple chunks/sub-processors consisting of a few pre/post-processors and a systolic array to accelerate dedicated operations. Specifically, the pre-processors include an accumulator array for performing column(token)-wise summation, and a divider array and a adder array for conducting element-wise divisions and additions, respectively; 
    In addition, the systolic array (SA) is partitioned into a smaller sub-array named {SA-Diag} to compute the matrix and diagonal matrix multiplications (i.e., $\mathbf{Q} \hat{\mathbf{k}}_{sum}^T$) considering their smaller number of multiplications, and a larger sub-array dubbed {SA-General} to process the remaining matrix multiplications (i.e., ${\mathbf{G}}=\Hat{\mathbf{K}}^T \mathbf{V}$ and ${\mathbf{Q}}{\mathbf{G}}$).}     
    \label{fig:micro-arch}
    \vspace{-0.5cm}
\end{figure*}
\vspace{0.5em}
\textbf{Opportunity $1$: Reduced Number of Multiplications and Low-Cost Pre/Post-Processing.}
As mentioned above, our proposed Taylor attention can largely reduce the number of expensive multiplications and further avoid the use of the hardware unfriendly softmax operation, while introducing a few low-cost pre/post-processing steps. As such, \textit{our Taylor attention in some sense trades higher-cost multiplications and softmax operation with lower-cost pre/post-processing steps}. Specifically, as formulated in Eq. (\ref{eq:mul}), our attention achieves $(\mathbf{n} / \mathbf{d})\times$ reduction in the number of multiplications, where the number of tokens $\mathbf{n}$ is in general much larger than that of the feature dimension $\mathbf{d}$ in ViTs, leading to $\mathbf{n} / \mathbf{d} >> 1$ (\textcolor{black}{e.g., $\mathbf{n} / \mathbf{d} \approx 3$ in DeiT models \cite{pmlr-v139-touvron21a}, and $12.25$, $3$, $1$ for the three stages in LeViT-128/128s \cite{Graham_2021_ICCV}, respectively}); WHile different from the three-step operations in the vanilla attention (see the bottom of Table \ref{tab:profiling}), our Taylor attention introduces several light-weight pre/post-processing steps to cooperate with the reduced number of matrix multiplications, i.e., as shown in Algorithm \ref{alg:vitality}, the Steps 1 and 3 for pre-processing the keys and values via column-wise accumulations and element-wise additions, respectively, the post-processing in Steps 4 and 5 via element-wise additions, and the post-processing in Step 6 via row-wise divisions. Note that although row-wise divisions are also used in the softmax operation of the vanilla attention, our Taylor attention reduces the number of divisions by $(\mathbf{n} / \mathbf{d})\times$ (see Eq. (\ref{eq:div})) as discussed above. 
Therefore, there exists an opportunity for the dedicated accelerator design to fully unleash the hardware efficiency benefits of our proposed Taylor attention's property of \textit{``trades higher-cost multiplications and softmax operation with lower-cost pre/post-processing steps"}. 

\vspace{0.5em}
\textbf{Opportunity $2$: Data Dependency across Different Steps.}
As summarized in Algorithm \ref{alg:vitality}, there exists data dependency in the sequentially processed steps of our Taylor attention. For example, 
1) the generated mean-centering key $\mathbf{\hat{K}}$ in Step $1$ serves as the input for computing both the global context matrix $\mathbf{G}$ in Step $2$ and the column sum of keys $\mathbf{\hat{k}_{sum}}$ in Step $3$;
2) Both the obtained Taylor denominator $\mathbf{t_{D}}$ in Step $4$ and the Taylor numerator $\mathbf{T_{N}}$ in Step $5$ are then used to compute the final Taylor attention score $\mathbf{Z}$ in Step $6$.
Such data dependency can cause large latency when accelerating the corresponding ViTs if not properly handled. For example, the computationally light pre/post-processing steps in our Taylor attention can account for about 50\% of the total ViT's profiling latency on the edge GPU, as shown in Table \ref{tab:profiling}. 
Hence, it is important for a dedicated accelerator to be equipped with proper pipeline designs to avoid the latency bottleneck due to the sequential execution pattern. 

\vspace{-0.2em}
\subsection{\textsc{ViTALiTy} Accelerator: Micro-Architecture}
\label{sec:micro-arch}

\textbf{Motivation.} As discussed in \emph{Opportunity $1$}, in addition to \emph{matrix multiplications}, our \textsc{ViTALiTy} accelerator is also expected to support \emph{column-wise accumulations, element-wise additions, and row-wise divisions} for pre/post-processing steps of the \textsc{ViTALiTy} algorithm. 
To achieve this goal, two typical designs can be considered: 
1) \emph{a single processor with reconfigurable processing units} to simultaneously support all the aforementioned four types of operations, where the key is to minimize the overhead of supporting the required reconfigurability; 
and 2) \emph{\textbf{a chunk-based accelerator} design integrating multiple chunks/sub-accelerators} with each dedicating to each type of operation, of which the advantage is that the reconfigurability overhead can be avoided but each chunk has a smaller amount of resources given the overall area constraint.
Considering that the pre/post-processing steps in our \textsc{ViTALiTy} algorithm contain mostly low-cost hardware efficient operations, we adopt the latter design with a larger chunk for computing the expensive multiplications and smaller chunks for handling other low-cost steps. 
\vspace{0.5em}

\textbf{Overview.} As shown in Fig.~\ref{fig:micro-arch}, our \textsc{ViTALiTy} accelerator adopts 1) a \emph{\textbf{four-level memory hierarchy}} including DRAM, SRAM, Network on Chip (NoC), and registers (Regs) within each computation unit to facilitate data reuses, and 2) a \emph{\textbf{multi-chunk design}} that integrates multiple dedicated chunks/sub-processors, including a few {pre/post-processors} and a {systolic array} for supporting the diverse operators in our Taylor attention.
In particular, the \emph{pre/post-processors} consist of an accumulator array, a divider array, and an adder array for performing column(token)-wise summation, element-wise divisions, and element-wise additions, respectively. Specifically, the \textcolor{light-blue}{\textbf{\emph{accumulator array}}} is to pre-process the keys and values for generating corresponding column summations via accumulating all elements along the column/token dimension, i.e., computing the column summation of the keys $\mathbf{1}_{n}^{T}\mathbf{K}$ (see Step $1$ in Algorithm \ref{alg:vitality}) and the column summation of both the mean-centering keys $\hat{\mathbf{k}}_{sum}$ and values ${\mathbf{v}}_{sum}$ (see Step $3$); 
The \textcolor{dark-orange}{\emph{\textbf{divider array}}} is to process element-wise divisions in our Taylor attention, i.e., dividing the elements in the column summation matrix $\mathbf{1}_{n}^{T}\mathbf{K}$ by the column/token dimension $n$ to obtain the column/token-wise mean matrix $\Hat{\mathbf{K}}$ (single-divisor division, see Step $1$ in Algorithm \ref{alg:vitality}) as well as conducting the division 
between the Taylor numerator matrix $T_{N}$ and the diagonal Taylor denominator matrix $t_{D}$ to generate the final Taylor attention score $\mathbf{Z}$ (multiple-divisors division, see Step $6$). As such, as illustrated in Fig.~\ref{fig:micro-arch} (see the upper left part), the divider array is designed to be reconfigurable for supporting both {\color{olive}{\ding{182}}} single-divisor division and {\color{dark-red}{\ding{183}}} multiple-divisors division patterns; 
In addition, the \textcolor{dark-green}{\emph{\textbf{adder array}}} is to perform the elements-wise additions/subtractions for obtaining the mean-centering keys $\Hat{\mathbf{K}}$ in Step $1$, the Taylor denominator $\mathbf{t}_{D}$ in Step $4$, and the Taylor numerator $\mathbf{T}_{N}$ in Step $5$. 

\begin{figure}[t]
    \centering
    \includegraphics[width=\linewidth]{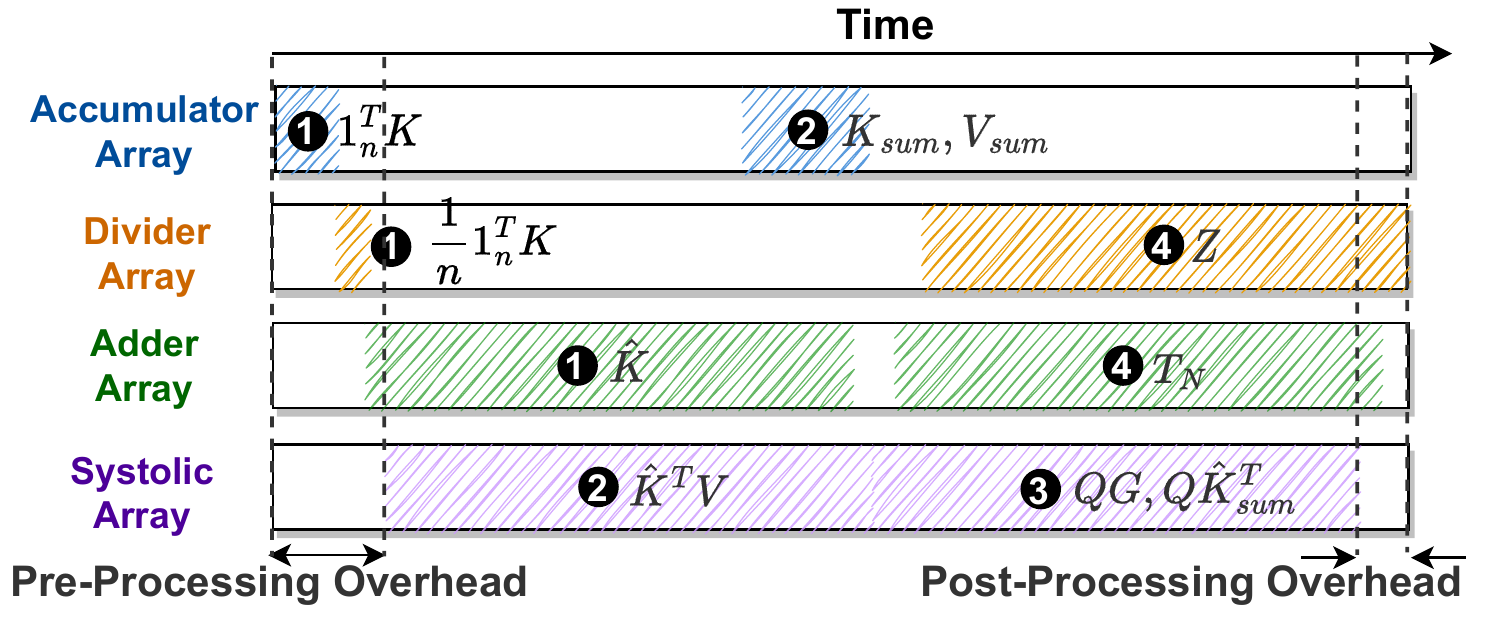}
    \vspace{-0.7cm}
    \caption{Illustrating the intra-layer pipeline design that minimizes both the pre- and post-processing overheads for enhancing the overall throughput.} 
    \label{fig:execution-steps}
    \vspace{-1.5em}
\end{figure}

Another important block in our \textsc{ViTALiTy} accelerator is the \textcolor{dark-purple}{\textbf{\emph{systolic array}}} which is to process matrix multiplications in our Taylor attention, i.e., $\mathbf{G} = \Hat{\mathbf{K}}^T \mathbf{V}$ in Step $2$, $\mathbf{Q} \hat{\mathbf{k}}_{sum}^T$ in Step $4$, and ${\mathbf{Q}}{\mathbf{G}}$ in Step $5$. 
Specifically, the \emph{systolic array} is partitioned into two parts donated as SA-Diag and SA-General, respectively, to compute $\mathbf{Q} \hat{\mathbf{k}}_{sum}^T$ and ${\mathbf{Q}}{\mathbf{G}}$ in parallel for 1) simultaneously calculating the Taylor denominator and numerator and thus can be then pipelined with the computation of the Taylor attention score in Step 6 to improve the overall throughput (see Section~\ref{sec:execution_steps} for details) and 2) reducing the data access cost of the queries $Q$ to improve the energy efficiency (see Section~\ref{sec:dataflow} for details).
As the number of multiplications in $\mathbf{Q} \hat{\mathbf{k}}_{sum}^T$ is much smaller than that in ${\mathbf{Q}}{\mathbf{G}}$, i.e., the former is $n\times d\times 1$ while the latter is $n\times d\times d$, it is natural to allocate fewer PEs (Processing Elements), i.e., only one PE column, for the SA-Diag sub-array.   
In addition to process matrix multiplications, the systolic array is also reused to compute the preceding linear projection and subsequent MLP module, considering their operations are similar.

\subsection{\textsc{ViTALiTy} Accelerator: Intra-Layer Pipeline Design}
\label{sec:execution_steps}
\begin{figure*} [!t]
    \centering
    \includegraphics[width=0.95\linewidth]{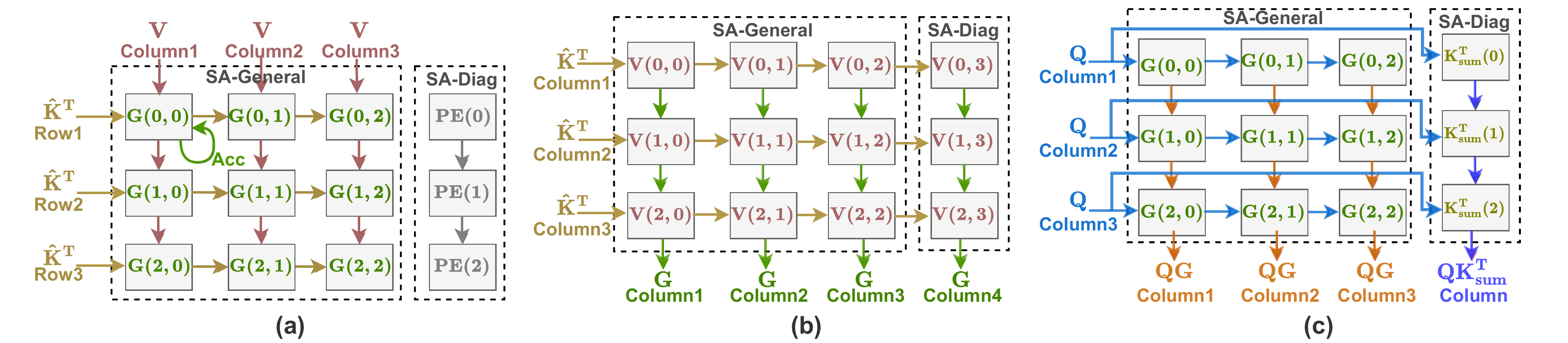}
    \vspace{-0.4cm}
    \caption{Illustrating two potential dataflows for \textsc{ViTALiTy} accelerator's systolic array: 1) \textit{G-stationary} adopts (a) output stationary to compute $\mathbf{G} = \Hat{\mathbf{K}}^T \mathbf{V}$ and (c) input stationary to process both $\mathbf{Q}\mathbf{G}$ and $\mathbf{Q}\Hat{\mathbf{K}}^T_{sum}$ for keeping $\mathbf{G}$ stationary within the PEs, and 2) \textit{down-forward accumulation dataflow} that adopts input stationary to handle all the multiplications (see (b) and (c)).  
    }
    \vspace{-1em}
    \label{fig:gs-df-dataflow}
\end{figure*}

As listed in Table \ref{tab:profiling}, general computing platforms, e.g., GPUs, cannot leverage the theoretical benefits of \textsc{ViTALiTy}'s Taylor attention for achieving actual hardware efficiency. One of the reasons is that the sequentially executed computation steps in our Taylor attention require dedicated pipeline design.
Considering the data dependency across different steps (see \emph{Opportunity $2$}) and our multi-chunk micro-architecture (see Section \ref{sec:micro-arch}), our \textsc{ViTALiTy} accelerator adopts an intra-layer pipeline design to enhance the overall throughput, as illustrated in Fig. \ref{fig:execution-steps} and discussed below:

\begin{enumerate} 
\vspace{-0.2em}
\item The accumulator and divider arrays first pre-process ${\mathbf{K}}$ to generate the column-wise mean of the keys $\overline{\mathbf{K}}$, where each element is then subtracted by the elements in the corresponding column of $\textbf{K}$ via the adder array to obtain the mean-centering keys $\hat{\mathbf{K}}$ (see Step $1$ of Algorithm \ref{alg:vitality}).

\item While computing $\hat{\mathbf{K}}$ via the adder array, the already generated ones are sent to both the \emph{systolic array} to multiply with ${\mathbf{V}}$ for computing the global context matrix ${\mathbf{G}}$ (i.e., Step $2$ of Algorithm \ref{alg:vitality}; see Fig. \ref{fig:gs-df-dataflow}(b) and Section \ref{sec:dataflow} for details) to reduce the pre-processing overhead, and the \emph{accumulator array} along with ${\mathbf{V}}$ to generate $\hat{\mathbf{k}}_{sum}$ and ${\mathbf{v}}_{sum}$ (see Step $3$) for decreasing accumulator array's idle time.

\item Once obtaining both $\mathbf{G}$ and ${\mathbf{k}}_{sum}^T$, the systolic array is reused and then partitioned into SA-General and SA-Diag to simultaneously process ${\mathbf{Q}\mathbf{G}}$ and $\mathbf{Q}\hat{\mathbf{k}}_{sum}^T$
with ${\mathbf{Q}}$ being broadcasted to both sub-arrays (see Fig. \ref{fig:gs-df-dataflow}(c) and Section \ref{sec:dataflow} for details). 

\item After calculating the first element in  $\mathbf{Q}\hat{\mathbf{k}}_{sum}^T$ and the first row of ${\mathbf{Q}}{\mathbf{G}}$, an extra adder unit and the adder array are used to compute the Taylor denominator $t_{D}$ and  numerator $\mathbf{T}_{N}$ in Steps $4$ and $5$, respectively, which are then processed by the divider arrray to generate the final Taylor attention score $\mathbf{Z}$ in Step $6$ of Algorithm \ref{alg:vitality}. This helps fuse the post-processing steps with matrix multiplications for reducing the post-processing overhead.
\end{enumerate} 

\begin{figure} [!t]
    \centering
    \includegraphics[width=\linewidth]{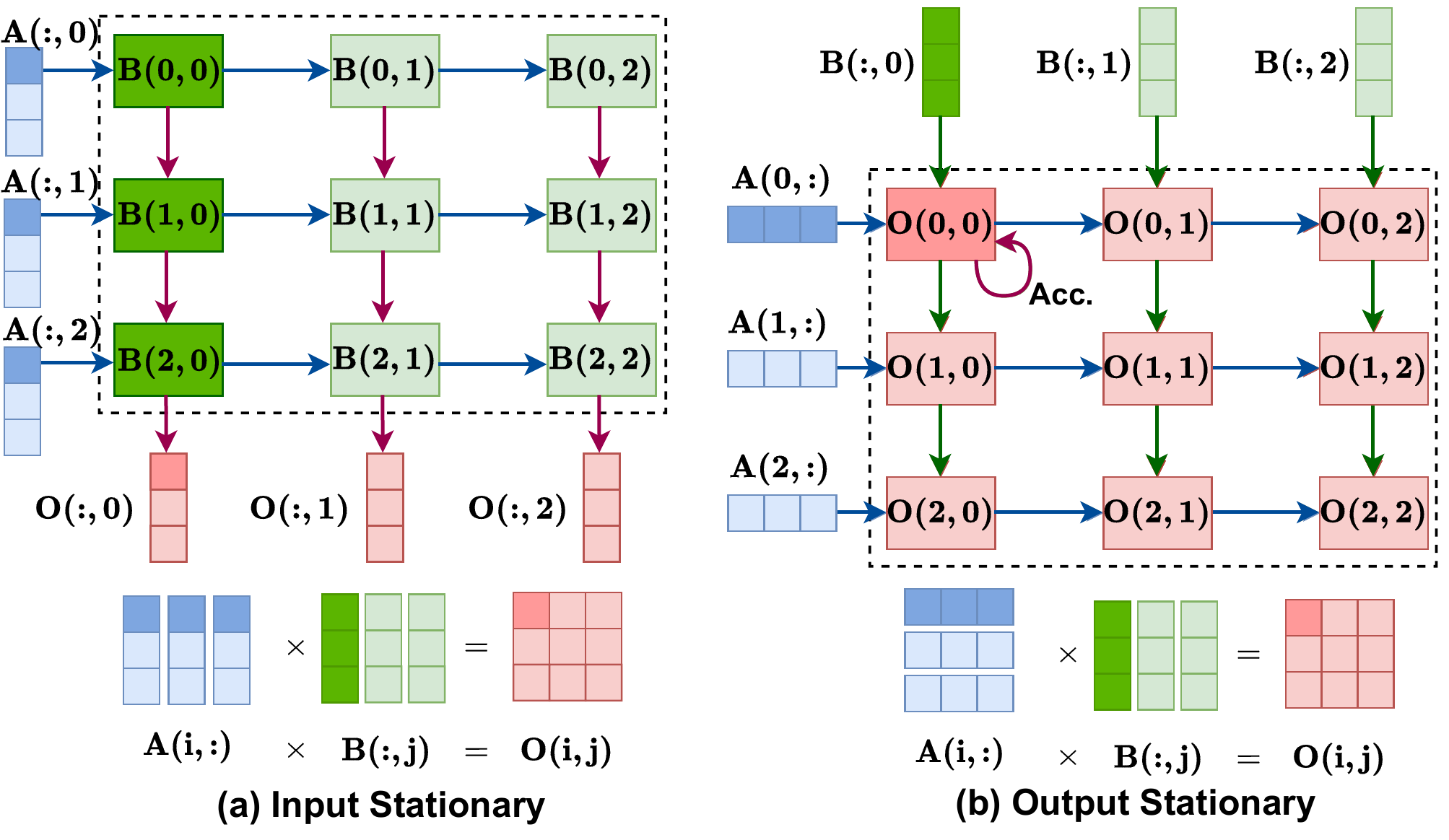}
    \vspace{-2em}
   \caption{Illustrating two typical dataflows for dense matrix multiplications: (a) input stationary and (b) output stationary, where the partial sums are down-forward accumulated in the former and inner-PE accumulated in the latter.} 
    \label{fig:is-os-dataflow}
    \vspace{-1em}
\end{figure}

\subsection{\textsc{ViTALiTy} Acc.: Down-Forward Accumulation Dataflow}
\label{sec:dataflow}

As matrix multiplication is the cost-dominant operation in our proposed Taylor attention, the adopted dataflow that determines how to temporally and spatially map multiplications onto the PE array is crucial to the achievable hardware efficiency of \textsc{ViTALiTy} accelerator. For better understanding, we first introduce two typical dataflows for processing dense matrix multiplications: \emph{input} and \emph{output stationary}.
As illustrated in Fig. \ref{fig:is-os-dataflow}, assuming $\mathbf{O}=\mathbf{A}\mathbf{B}$, we can see that
1) for input stationary, $\mathbf{B}$ stays stationary in the PE array and each column of $\mathbf{A}$ is horizontally traversed to one PE row, where each PE column is responsible for processing one vector dot-product between one row of $\mathbf{A}$ and one column of $\mathbf{B}$ and the partial sums are then down-forward accumulated to gather the final outputs at the bottom-most PEs (\textbf{{down-forward accumulation}});
2) For output stationary, each row of $\mathbf{A}$ instead is horizontally sent to one PE row while each column of $\mathbf{B}$ is vertically mapped to one PE column, where each PE computes the vector dot-product between one row of $\mathbf{A}$ and one column of $\mathbf{B}$ and the partial sums are then temporally accumulated within the PEs (\textbf{{inner-PE accumulation}}).

\vspace{-0.1em}
As discussed in Section~\ref{sec:execution_steps}, our systolic array is leveraged to first perform $\mathbf{G} = \Hat{\mathbf{K}}^T \mathbf{V}$ and then reused and partitioned into two sub-blocks dubbed SA-General and SA-Diag to simultaneously compute $\mathbf{Q}\mathbf{G}$ and $\mathbf{Q}\Hat{\mathbf{k}}^T_{sum}$, respectively.
Considering the two typical dataflows above for one single dense matrix multiplication, there exist two potential dataflows for handling the consecutive matrix multiplications in our Taylor attention (i.e., $\mathbf{G} = \Hat{\mathbf{K}}^T \mathbf{V}$, $\mathbf{Q}\mathbf{G}$, and $\mathbf{Q}\Hat{\mathbf{k}}^T_{sum}$ in Algorithm \ref{alg:vitality}): \emph{G-stationary} and \emph{down-forward accumulation dataflows}.

Specifically, 1) as $\mathbf{G}$ serves as both the output of $\mathbf{G} = \Hat{\mathbf{K}}^T \mathbf{V}$ and the input of $\mathbf{Q}\mathbf{G}$, \textcolor{black}{\textbf{{G-stationary}} is the most intuitive dataflow that adopts output-stationary for computing $\mathbf{G} = \Hat{\mathbf{K}}^T \mathbf{V}$ and then keeps $\mathbf{G}$ stationary within the PEs to serve as the input for computing $\mathbf{Q}\mathbf{G}$ via input-stationary. As such, the PEs need to be reconfigurable for simultaneously supporting both inner-PE accumulation for output stationary (see Fig. \ref{fig:gs-df-dataflow} (a)) and down-forward accumulation for input stationary (see Fig. \ref{fig:gs-df-dataflow} (c))}; 2) Another alternative is the \textbf{{down-forward accumulation dataflow}} that adopts input-stationary for computing all the matrix multiplications in our attention, i.e., $\mathbf{V}$ stays stationary when computing $\Hat{\mathbf{K}}^T \mathbf{V}$ (see Fig. \ref{fig:gs-df-dataflow} (b)) while $\mathbf{G}$ and $\Hat{\mathbf{k}}^T_{sum}$ remain stationary in the SA-General and SA-Diag sub-blocks, respectively, with $\mathbf{Q}$ being broadcasted to process $\mathbf{Q}\mathbf{G}$ and $\mathbf{Q}\Hat{\mathbf{k}}^T_{sum}$ in parallel (see Fig. \ref{fig:gs-df-dataflow} (c)). Hence, the former enhances data locality of $\mathbf{G}$ for minimizing its data access cost but requires overhead to reconfigure the PEs for simultaneously supporting the above two accumulation patterns, while the latter simplifies the PE design at a cost of increased data access cost. As the energy consumed by the systolic array dominates the overall energy cost instead of the data access cost from SRAM (see Table \ref{tab:energy_comparison}), we adopts the down-forward accumulation dataflow for boosting the overall energy efficiency (see ablation study in Section \ref{sec:ablation}).

%% file: results.tex
\section{Evaluation and Analysis}
\subsection{Experiment Setting}
\textbf{Models and Methods.} We evaluate on popular ViT models, including 1) vanilla ViTs - DeiT-Base/Small/Tiny~\cite{pmlr-v139-touvron21a}, 2) lightweight ViT models - MobileViT-xxs/xs~\cite{mehta2021mobilevit}, and 3) hybrid ViT models - LeViT-128s/128~\cite{Graham_2021_ICCV}.
Various methods are benchmarked with: \textsc{Baseline} (ViTs with vanilla softmax attentions), \textsc{Sparse} (Sanger \cite{sanger} with a sparsity threshold of $T=0.02$), \textsc{LowRank} (ViTs with Taylor attentions on pre-trained models), and \textsc{ViTALiTy} (trained using both low-rank and sparse attentions, but inference with only low-rank attentions). While Sanger is originally designed for NLP tasks, we implement its algorithm to train the ViT models. Model accuracy is reported on the ImageNet\cite{deng2009imagenet} validation dataset. 

\textbf{Hardware Settings.}
To verify the effectiveness of our \textsc{ViTALiTy} accelerator, we consider three general computing hardware platforms, including 1) CPU (Intel(R) Xeon(R) Gold 6230), 2) Edge GPU (NVIDIA Tegra X2), and 3) GPU (NVIDIA 2080Ti), and 4) a SOTA dedicated attention accelerator Sanger \cite{sanger}. 
Note that for fair comparisons, we scale up \textsc{ViTALiTy}'s hardware resource to be comparable with the aforementioned baselines. Specifically, when benchmarking with the general computing platforms, we follow \cite{qu2022dota} to scale up \textsc{ViTALiTy}'s accelerator to have a comparable peak throughput as that of the platform, \textcolor{black}{and for benchmarking with \textsc{Sparse} (Sanger), we adopt the comparable hardware budgets as Sanger.} 
We compare our dedicated accelerator over these four baselines in terms of both energy efficiency and speedup. 

\subsection{Implementation}
\begin{table}[t]
\centering
\caption{{Configurations of \textsc{ViTALiTy} and Sanger \cite{sanger} accelerators.}}
\vspace{-0.5em}
\setlength{\tabcolsep}{0.5mm}{
{
\begin{tabular}{|cc|c|c|c|}
\hline 
\multicolumn{1}{|c|}{$\textsc{\textbf{ViTALiTy}}$}                       & \textbf{Component}         & \textbf{Parameter}            & \begin{tabular}[c]{@{}c@{}}\textbf{Area}\\ \textbf{($mm^{2}$)}\end{tabular} & \begin{tabular}[c]{@{}c@{}}\textbf{Power}\\ \textbf{(mW)}\end{tabular} \\ \hline
\multicolumn{1}{|c|}{\multirow{3}{*}{\begin{tabular}[c]{@{}c@{}}Pre/Post- \\ Processors\end{tabular}}} & Accumulator Array & $64\times1$ $16$-bit  & $0.209$                                                                      & $92.83$                                                \\
\multicolumn{1}{|c|}{}                                                                                   & Adder Array       & $64\times1$ $16$-bit  & $0.012$                                                                      & $6.34$                                                 \\
\multicolumn{1}{|c|}{}                                                                                   & Divider Array     & $64\times1$ $16$-bit  & $0.562$                                                                      & $46.26$                                                \\ \hline
\multicolumn{1}{|c|}{\multirow{2}{*}{\begin{tabular}[c]{@{}c@{}}Systolic \\ Array\end{tabular}}}       & SA-General        & $64\times64$ $16$-bit & $3.595$                                                                      & $1277$                                                 \\
\multicolumn{1}{|c|}{}                                                                                   & SA-Diag           & $64\times1$ $16$-bit  & $0.053$                                                                      & $15.18$                                                \\ \hline
\multicolumn{1}{|c|}{Memory}                                                                             & {[}$\mathbf{Q}$, $\mathbf{K}$, $\mathbf{V}$, $\mathbf{O}${]}  & $50$ KB$\times4$      & $0.792$                                                                      & $22.9$                                                 \\ \hline
\multicolumn{2}{|c|}{$\textbf{Overall}$}                                                                                                & $28$ nm                 & $\mathbf{5.223}$                                                                     & $\mathbf{1460}$                                                \\ \hline \hline
\multicolumn{1}{|c|}{$\textsc{\textbf{Sanger}}$}                                                   & \textbf{Component}         & \textbf{Parameter}            & \begin{tabular}[c]{@{}c@{}}\textbf{Area}\\ ($mm^{2}$)\end{tabular} &

\begin{tabular}[c]{@{}c@{}}\textbf{Power}\\ \textbf{(mW)}\end{tabular} \\ \hline
\multicolumn{1}{|c|}{\multirow{3}{*}{\begin{tabular}[c]{@{}c@{}}Pre/Post-\\ Processors\end{tabular}}} & Pre-Processor     & 64 $\times$ $64$ $44$-bit  & $0.430$                                                                      & $182.8$                                               \\
\multicolumn{1}{|c|}{}                                                                                   & Pack \& Split     & $64\times64$ $1$-bit   & $0.016$                                                                      & $0.64$                                                 \\
\multicolumn{1}{|c|}{}                                                                                   & Divider Array     & $64\times1$ $16$-bit  & $0.562$                                                                      & $46.26$                                                \\ \hline
\multicolumn{1}{|c|}{\multirow{2}{*}{\begin{tabular}[c]{@{}c@{}}Systolic \\ Array\end{tabular}}}       & RePE              & $64\times16$ $16$-bit  & \multirow{2}{*}{$3.393$}                                                     & \multirow{2}{*}{$1198.35$}                             \\
\multicolumn{1}{|c|}{}                                                                                   & EXP               & $64\times1$ $16$-bit  &                                                                            &                                                      \\ \hline
\multicolumn{1}{|c|}{Memory}                                                                             & {[}$\mathbf{Q}$, $\mathbf{K}$, $\mathbf{V}$, $\mathbf{O}${]}  & $50$ KB $\times4$      & $0.792$                                                                      & 22.90                                                \\ \hline
\multicolumn{2}{|c|}{$\textbf{Overall}$}                                                                                                & $28$ nm                 & $\mathbf{5.194}$                                                                     & $\mathbf{1450}$                                                \\ \hline
\end{tabular}}} \label{table:hardware_config}
\vspace{-0.5cm}
\end{table}
\textbf{Software Implementation.} We fine-tune the pre-trained ViT models from \cite{rw2019timm}, and follow the training recipe in \cite{pmlr-v139-touvron21a}. For inference, we implement Algorithm~\ref{alg:vitality} for DeiT, MobileViT and LeViT models. To further boost accuracy, we apply token based knowledge distillation \cite{pmlr-v139-touvron21a} during training.

\textbf{Hardware Implementation.}
To evaluate the performance of \textsc{ViTALiTy}, we implement a cycle-accurate simulator for our dedicated accelerator to obtain fast and reliable estimations, which are verified against the RTL implementation to ensure the correctness. The adopted unit energy and area are synthesized on a $28$ nm CMOS technology using Synopsys tools (e.g., Design Compiler for gate-level netlist \cite{DC}) at a frequency of $500$ MHz. For a fair comparison, we also implement a cycle-accurate simulator for the baseline accelerator, Sanger \cite{sanger}, and compare the simulated results with the reported performance in the Sanger \cite{sanger} paper to ensure the correctness. {As shown in Table \ref{table:hardware_config}, we evaluate both \textsc{ViTALiTy} accelerator and Sanger accelerator under a comparable area and power when being synthesized under the same CMOS technology and clock frequency.} 

\begin{figure}
    \centering
    \includegraphics[width=\linewidth]{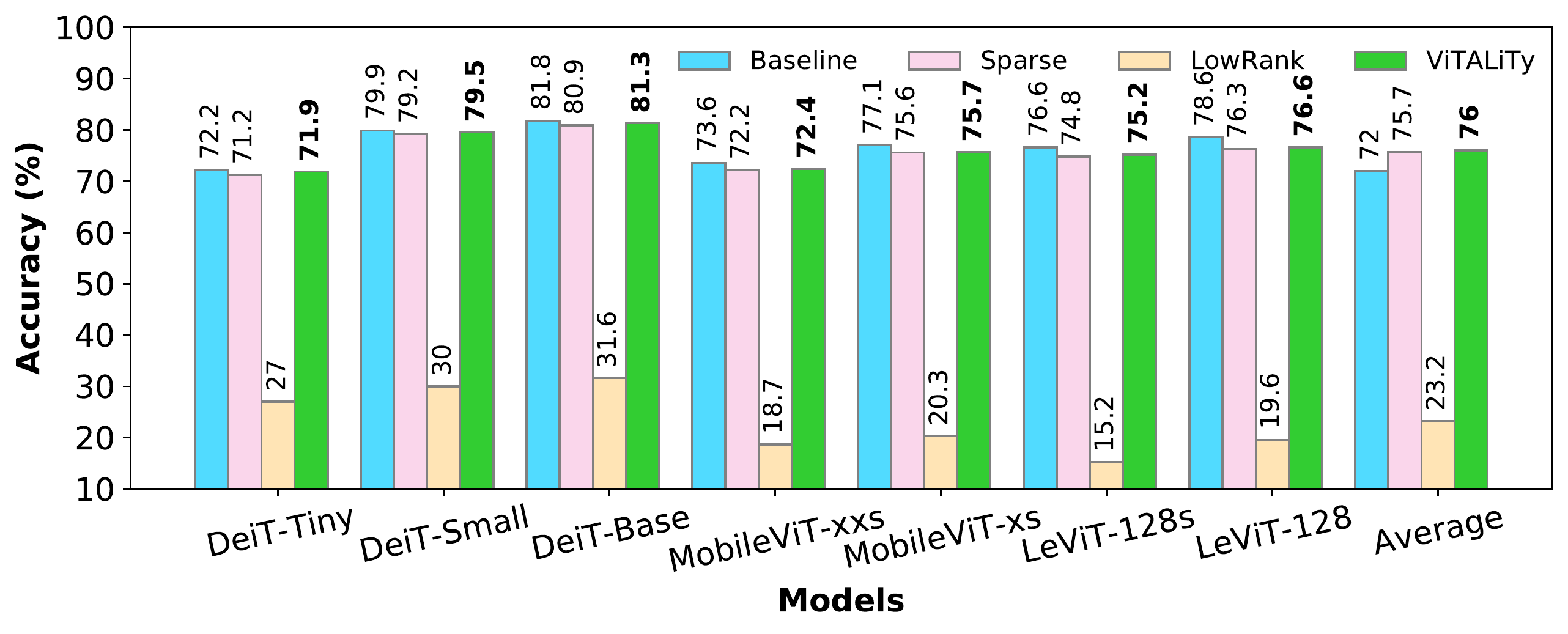}
    \vspace{-2em}
    \caption{{Accuracy comparison between \textsc{ViTALiTy} and other methods across various ViT models. Here, \textsc{Baseline} is vanilla softmax attention, \textsc{Sparse} is Sanger~\cite{sanger}, and \textsc{LowRank} refers to applying linear Taylor attention on pre-trained model. \textsc{ViTALiTy} outperforms both \textsc{Sparse} and \textsc{LowRank}.}}
    \vspace{-1em}
    \label{fig:baseline}
\end{figure}
\begin{figure}[t]
    \centering
    \includegraphics[width=\linewidth]{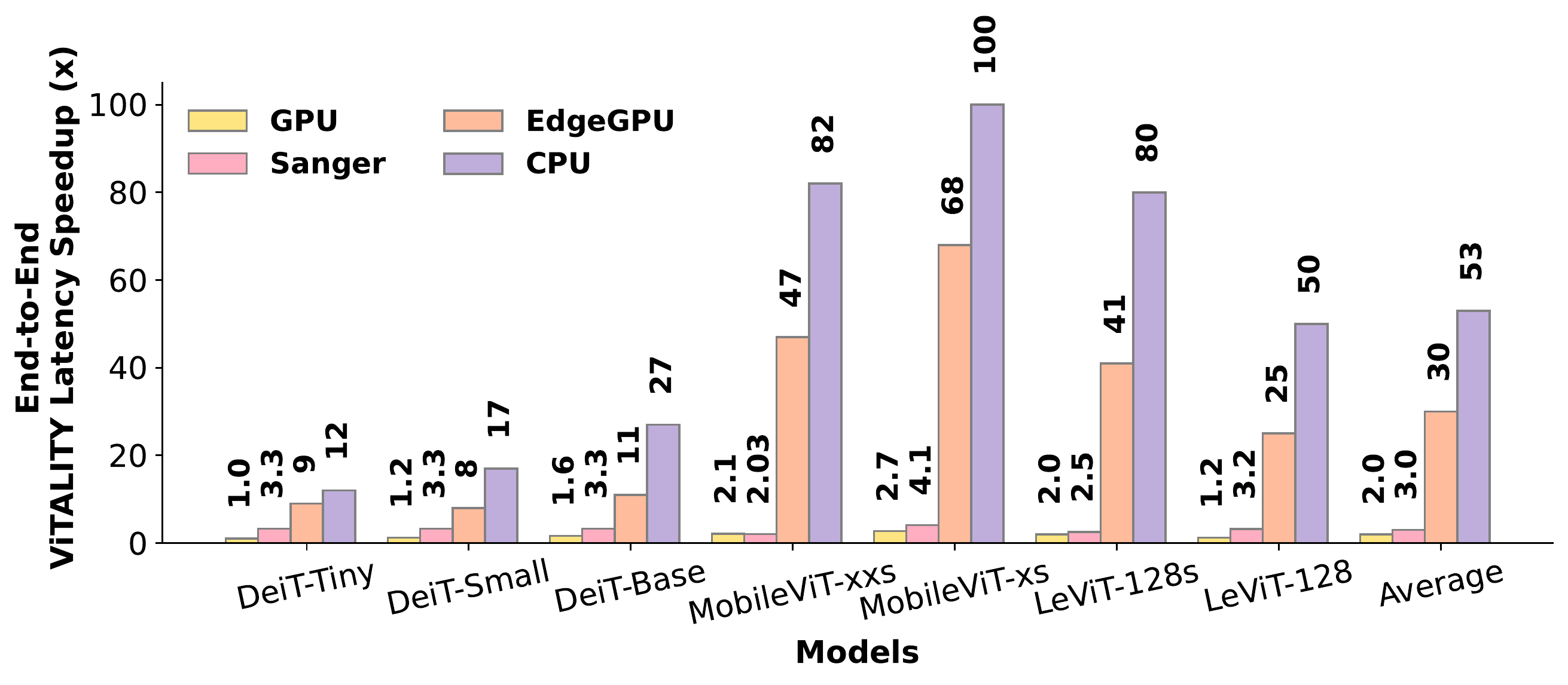}
    \vspace{-2em}
    \caption{{End-to-End Latency Speedup of  \textsc{ViTALiTY} accelerator.}}
    \vspace{-1em}
    \label{fig:speedupLatency}
\end{figure}

\begin{table}[t]
\centering
\caption{Accuracy vs FLOPS (Attention) tradeoff for various methods.}
\vspace{-0.5em}
\begin{tabular}{|c|c|c|c|}
\hline
\textbf{Method }& \textbf{Type} & \textbf{Accuracy} (\%) & \textbf{FLOPs} (G) \\ \hline
\textsc{Baseline} & Quadratic & $72.2$ & $0.50$ \\ \hline
\textsc{ViTALiTy} (ours) & \textbf{Linear} & $\textbf{71.9}$ & $\textbf{0.33}$ \\ \hline
Linformer~\cite{linformer} & Linear & $69.5$ & $0.35$ \\ \hline
Performer~\cite{choromanski2021rethinking} & Linear & $68.3$ & $0.40$ \\ \hline
\textsc{Sanger}~\cite{sanger} & Sparse & $71.2$ & $0.33$ \\ \hline
SViTE~\cite{chen2021chasing} & Sparse & $71.7$ & $0.38$ \\ \hline
UVC~\cite{yu2022unified} & Sparse & $71.8$ & $0.30$ \\ \hline
\end{tabular}
\label{tab:accFLOPS}
\vspace{-1em}
\end{table}

\subsection{Performance Analysis}
Here, we analyze the performance of the proposed \textsc{ViTALiTy} in terms of accuracy, latency, and energy efficiency. 

\textbf{Accuracy.}
In Fig.~\ref{fig:baseline}, we demonstrate the accuracy achieved by various methods. We observe that training the various ViT models with \textsc{ViTALiTy} consistently \textbf{improves} the model accuracy with $\textbf{45\%}-\textbf{60\%}$ on the \textsc{LowRank} method, and with $\textbf{0.1\%}-\textbf{0.7\%}$ on the \textsc{Sparse} method. This demonstrates that the proposed \textsc{ViTALiTy} is a version of Scatterbrain~\cite{chen2021scatterbrain} where the combination of low-rank and sparse approximation training achieves better accuracy than the individual components. Once the model is trained, in contrast to Scatterbrain~\cite{chen2021scatterbrain} and \textsc{Sparse}, \textsc{ViTALiTy} simply employs the low-rank component for inference by computing the linear Taylor Attention to achieve this improved accuracy without the runtime overhead of sparse attention. It is worth recalling that \textsc{ViTALiTy} uses sparse approximation for capturing strong connections during training compared to vanilla softmax attention in \textsc{Baseline}. Hence, the accuracy of \textsc{ViTALiTy} falls short of \textsc{Baseline} accuracy by $0.3\%-2\%$ as expected. We trade the above accuracy drop of \textsc{ViTALiTy} trained models with the latency speedup and energy efficiency achieved using linear attention and dedicated accelerator for inference compared to running \textsc{Baseline} method with full softmax attention incurring quadratic costs. {Table \ref{tab:accFLOPS} shows that \textsc{ViTALiTy} outperforms other linear and sparse attention with better or comparable (UVC) accuracy versus FLOPs (attention) tradeoff.}
\vspace{0.5em}

\textbf{Latency Speedup.}
Fig. \ref{fig:speedupLatency} shows latency speedup of our \textsc{ViTALiTy} accelerator where it consistently outperforms all the aforementioned hardware baselines, validating the effectiveness of our proposed linear Taylor Attention algorithm on reducing the computation complexity of attention, and proposed intra-layer pipeline design of the accelerator in boosting the overall throughput. Specifically, on benchmarking the acceleration of core attention on general platforms, our \textsc{ViTALiTy} accelerator achieves an average $\textbf{236}\times$, $\textbf{239}\times$, and $\textbf{9}\times$ speedup over CPU, Edge GPU, and GPU, respectively. When comparing the end-to-end latency, our \textsc{ViTALiTy} accelerator is $\textbf{53}\times$, $\textbf{30}\times$, and $\textbf{2}\times$ faster over CPU, Edge GPU, and GPU, respectively. Furthermore, compared to Sanger \cite{sanger}, \textsc{ViTALiTy} gains an average $\textbf{7}\times$ speedup on the attention acceleration with $\textbf{3}\times$ speedup on end-to-end acceleration. Additionally, we verify the effectiveness of \textsc{ViTALiTy} accelerator by comparing it with the accelerator SALO \cite{SALO} which is designed for LongFormer \cite{Longformer} linear attention. It enables hybrid sparse attention mechanisms including sliding window attention, dilated window attention, and global attention. When comparing both accelerators under the same hardware budget for DeiT-Tiny and DeiT-Small models, \textsc{ViTALiTy} achieves up to \textbf{4.7$\times$} and \textbf{5.0$\times$} speedup on the attention acceleration,  respectively, under a comparable accuracy.

\begin{figure}[t]
    \centering
    \includegraphics[width=0.9\linewidth]{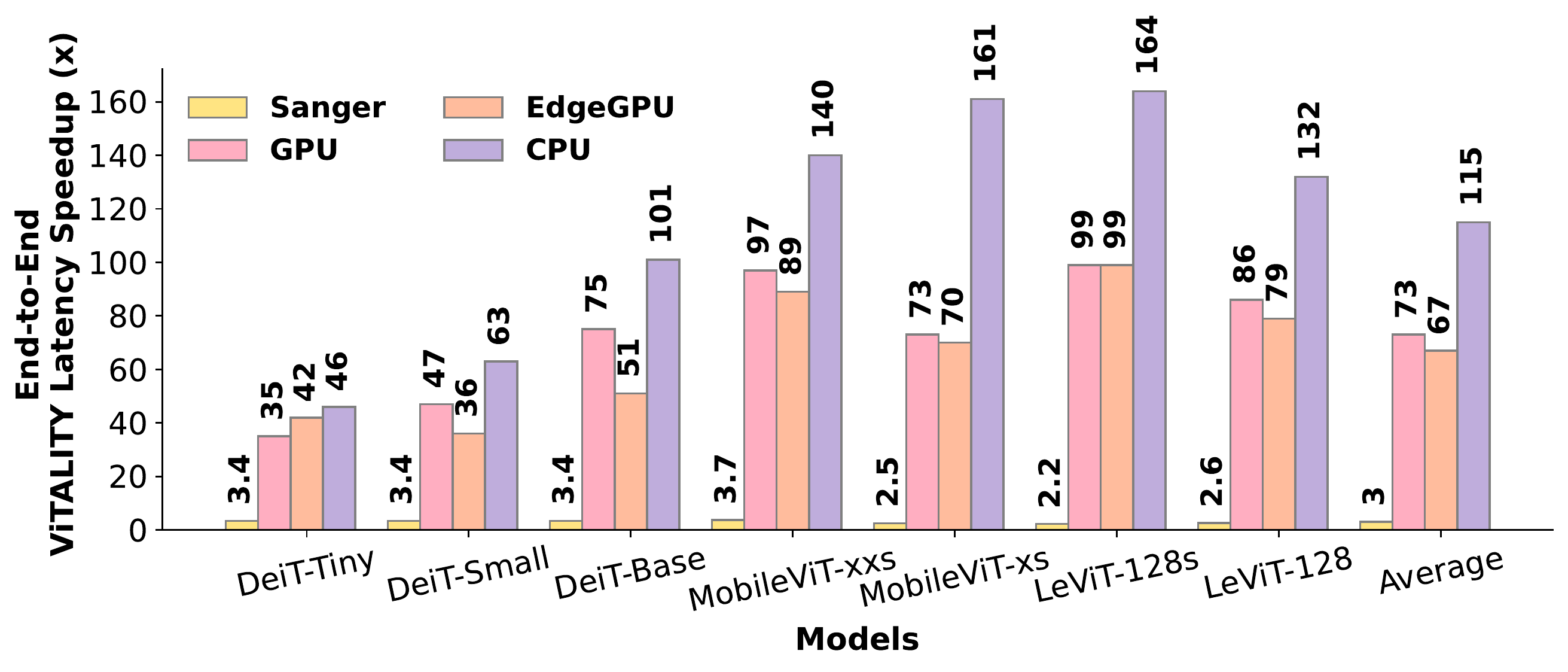}
    \vspace{-1em}
    \caption{{Energy Efficiency comparison of \textsc{ViTALiTY} accelerator.}}
    \label{fig:EnergyEff}
    \vspace{-1em}
\end{figure}
\begin{figure}[t]
    \centering
    \includegraphics[width=0.9\linewidth]{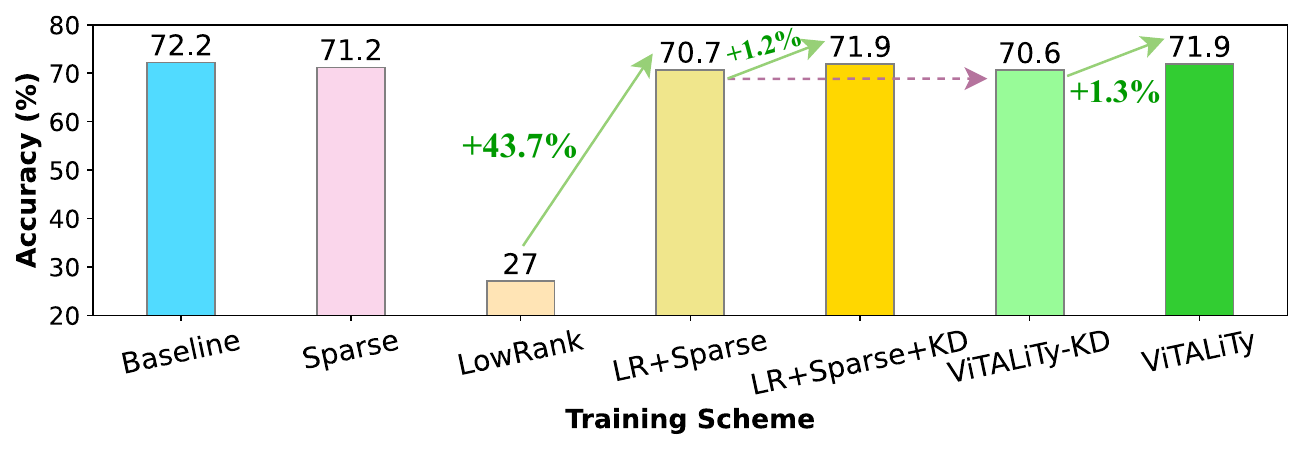}
    \vspace{-1em}
    \caption{{Ablation study of \textsc{ViTALiTy} on DeiT-Tiny model. LR denotes \textsc{LowRank}, and KD denotes Knowledge Distillation.}}
    \label{fig:ablation}
    \vspace{-1em}
\end{figure}
\vspace{0.5em}

\textbf{Energy Efficiency.}
Fig.\ref{fig:EnergyEff} shows the \textsc{ViTALiTy} accelerator's improvement in energy efficiency over other hardware baselines, demonstrating the advantage of its co-design framework. 
For core attention steps, 
our \textsc{ViTALiTy} accelerator achieves an average $\textbf{537}\times$, $\textbf{309}\times$, $\textbf{187}\times$, and $\textbf{6}\times$ better energy efficiency , and for the end-to-end performance, it offers an average $\textbf{115}\times$, $\textbf{67}\times$, $\textbf{73}\times$, and $\textbf{3}\times$ better energy efficiency, over CPU, Edge GPU, GPU, and Sanger \cite{sanger}, respectively.

\subsection{Ablation Study}
\label{sec:ablation}
Here, we discuss the ablation study of \textsc{ViTALiTy} training scheme on the DeiT-Tiny model, the effect of sparsity threshold on the model accuracy, and dataflow in accelerator design.

\textbf{Training Scheme.}
Fig.~\ref{fig:ablation} demonstrates the accuracy obtained by various training schemes as part of our ablation study on the proposed \textsc{ViTaLiTy} method. We recall that the \textsc{ViTaLiTy} method involves training ViT by unifying linear Taylor Attention as \textsc{LowRank} component, and the Sanger~\cite{sanger} induced sparsity mask with threshold, $T=0.5$ as the \textsc{Sparse} component. We also incorporate token-based knowledge distillation (KD). For inference, \textsc{ViTaLiTy} simply uses the linear Taylor Attention without any additional sparse component. Compared to \textsc{Baseline} accuracy $\textbf{72.2\%}$ with softmax attention, we observe that \textsc{Sparse} (Sanger~\cite{sanger}, $T=0.02$) achieves $\textbf{71.2\%}$ accuracy.

\vspace{0.5em}

\noindent
\textcolor{Orange}{\Pointinghand \textbf{ \underline{\textsc{Sparse} boosts the accuracy of \textsc{LowRank}}}}

\textsc{LowRank} method uses linear Taylor Attention as drop-in replacement of softmax attention in pre-trained ViT models for inference and suffers from poor model accuracy of $\textbf{27\%}$. This supports our claim in Section~\ref{sec:TA} that it is unlikely all (query,key) pairs have similarity within [-1,1). On unifying the above linear Taylor Attention (low-rank) and Sanger~\cite{sanger} sparsity with $T=0.5$ as \textsc{LowRank+Sparse} for fine-tuning the model and in inference, we observe improved accuracy of $\textbf{70.7\%}$ (with boost of $43.7\%$ over \textsc{LowRank} method). This demonstrates that the model now successfully captures the strong connections using the \textsc{Sparse} component which was missing using just the linear Taylor Attention from the \textsc{LowRank} method. Applying KD during fine-tuning of model additionally improves the accuracy by $1.2\%$ to $\textbf{71.9\%}$.
\vspace{0.5em}

\noindent
\textcolor{orange}{\Pointinghand \textbf{ \underline{\textsc{Sparse} is not required during inference}}}

\begin{wrapfigure}{r}{0.2\textwidth}
    \centering
    \vspace{-1em}
    \includegraphics[width=0.8\linewidth]{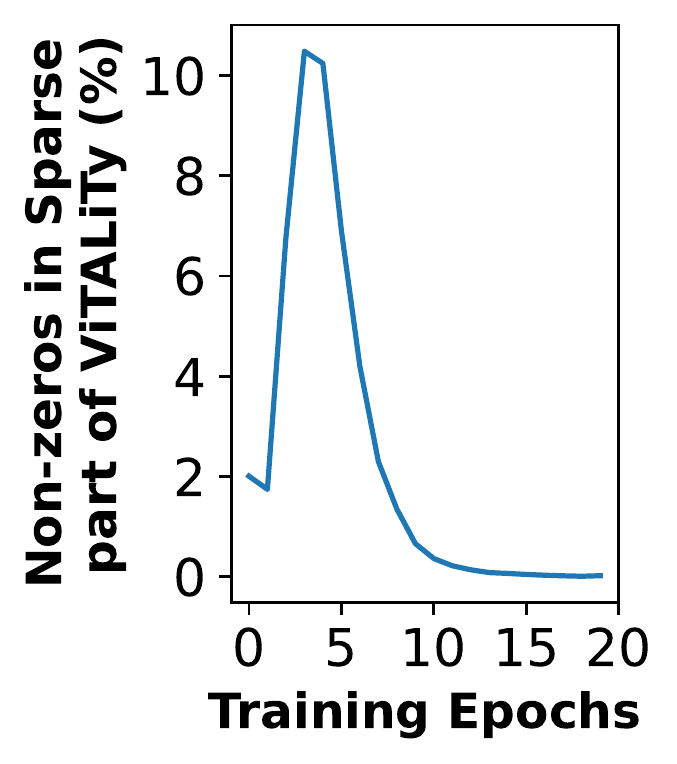}
    \vspace{-1em}
    \caption{{Sparse component vanishes after $10$ epochs, hence, can be dropped during inference.}}
    \label{fig:dropSparse}
\end{wrapfigure}

Fig.~\ref{fig:dropSparse} shows that the sparse component in \textsc{ViTALiTy-KD} vanishes after boosting the accuracy in initial training epochs of DeiT-Tiny. Hence, it can be dropped during inference with negligible effect on model accuracy after fine-tuning the model with unified \textsc{LowRank+Sparse}. This critical observation implies that the \textsc{Sparse} acts as a regularizer for generalizing the linear Taylor Attention such that during inference, the corresponding dense \textsc{LowRank} component is self-sufficient. This empowers the proposed \textsc{ViTALiTy} method to deploy ViT models on edge devices by avoiding the overhead of runtime sparsity computation during inference for various inputs. Applying KD further boosts the accuracy of \textsc{ViTALiTy} by $1.3\%$ to $\textbf{71.9\%}$ which matches the \textsc{LowRank+Sparse+KD} accuracy attained by additionally computing the sparse component during inference.
\vspace{0.5em}

\begin{wrapfigure}{r}{0.3\textwidth}
    \centering
    \includegraphics[width=0.3\textwidth]{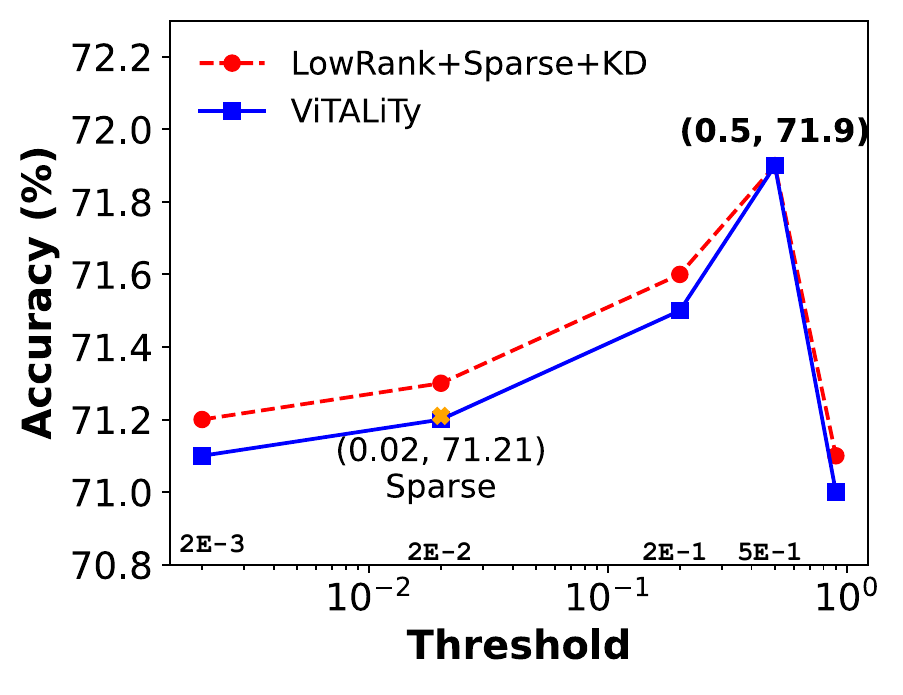}
    \vspace{-0.8cm}
    \caption{{Effect of sparsity threshold on DeiT-Tiny model accuracy. Optimal $T=0.5$.}}
    \vspace{-0.3cm}
    \label{fig:threshold}
\end{wrapfigure}

\textbf{Sparsity Threshold.}
Fig.~\ref{fig:threshold} illustrates the effect of sparsity thresholds [$0.002, 0.02, 0.2, 0.5, 0.9$] on DeiT-Tiny model accuracy. The \textsc{Sparse} method achieves accuracy of $71.2\%$ (Fig.~\ref{fig:ablation}) using the default threshold, $T=0.02$ defined in Sanger~\cite{lu2021sanger}. By incorporating \textsc{LowRank+Sparse+KD}, the model achieves slightly improved accuracy of $71.3\%$ at same threshold value, whereas, by using \textsc{ViTALiTy} which drops the sparse component during inference, the model retains the same accuracy of $71.2\%$ as the \textsc{Sparse} method.
\vspace{0.5em}

\noindent
\textcolor{orange}{\Pointinghand \textbf{ \underline{\textsc{LowRank} renders \textsc{Sparse} to exhibit high sparsity}}}

As the sparsity threshold increases, more zeros are introduced in the sparse attention matrix generated by applying Sanger sparsity masks on quantized softmax attention. This reduces the contribution of sparse component relative to the low-rank component (linear Taylor Attention) in \textsc{LowRank+Sparse+KD}, and \textsc{ViTALiTy} methods. When $T=1$, the sparse component vanishes completely during training and inference which leaves only the low-rank component, thereby, signalling accuracy drop when $T=0.9$ as demonstrated in Fig.~\ref{fig:threshold}. 
For low threshold values, there is low sparsity which implies large contribution of softmax attention which overpowers the low-rank properties of linear Taylor Attention. As sparsity threshold increases, \textsc{LowRank} component becomes more prominent as the approximation of softmax attention leads to highly sparse matrix. This helps improve the model accuracy for both methods (Fig.~\ref{fig:threshold}). We observe the optimal threshold value is empirically achieved at $T=0.5$, where, the low-rank properties of linear Taylor Attention in \textsc{ViTALiTy} renders the sparse attention matrix to exhibit high sparsity during training which yields the best accuracy of $71.9\%$. At this threshold, \textsc{ViTALiTy} with no sparse component during inference matches the accuracy of \textsc{LowRank+Sparse+KD}.

\textbf{Dataflow in Systolic Array.}
\begin{table}[]
\centering
\caption{Energy comparison of G-stationary and proposed down-forward accumulation dataflows for Taylor Attention}
\setlength{\tabcolsep}{0.3em}
\resizebox{\linewidth}{!}{
\begin{tabular}{|c|cc|cc|cc|cc|cc|}
\hline
\multirow{2}{*}{\begin{tabular}[c]{@{}c@{}}\textbf{Energy}\\ ($10^{-6}J$)\end{tabular}} & \multicolumn{2}{c|}{\textbf{DeiT-Base}}  & \multicolumn{2}{c|}{\textbf{MobileViT-xxs}} & \multicolumn{2}{c|}{\textbf{MobileViT-xs}} & \multicolumn{2}{c|}{\textbf{LeViT-128s}} & \multicolumn{2}{c|}{\textbf{LeViT-128}}  \\ \cline{2-11} 
                                                                               & \textbf{GS}            & \textbf{Ours}            & \textbf{GS}               & \textbf{Ours}            & \textbf{GS}              & \textbf{Ours}            & \textbf{GS}             & \textbf{Ours}           & \textbf{GS}            & \textbf{Ours}           \\ \hline
Data Access                                                                    & $\textbf{2.92}$ & $3.76$            & $\textbf{0.12}$    & $0.15$            & $\textbf{0.23}$   & $0.28$            & $\textbf{0.09}$  & $0.12$           & $\textbf{0.14}$ & $0.19$           \\ \hline
Other Processors                                                               & $3.92$          & $3.92$            & $0.15$             & $0.15$            & $0.23$            & $0.23$            & $0.12$           & $0.12$           & $0.19$          & $0.19$           \\ \hline
Systolic Array                                                                 & $215$           & $\textbf{191}$    & $13.2$             & $\textbf{10.3}$   & $23.3$            & $\textbf{20.1}$   & $11.1$           & \textbf{9.03}  & $16.6$          & $\textbf{13.3}$  \\ \hline
\textbf{Overall}                                                                        & $222$           & $\textbf{198}$   & $13.5$             & $\textbf{10.6}$   & $23.8$            & $\textbf{20.6}$   & $11.3$           & $\textbf{9.27}$  & $16.9$          & $\textbf{13.7}$  \\ \hline
\end{tabular}} \label{tab:energy_comparison}
\end{table}
In Table \ref{tab:energy_comparison}, we observe that the
(1) overall energy consumption of our proposed down-forward accumulation dataflow is consistently lower than the G-Stationary (GS) dataflow when benchmarking Taylor Attention on various ViT models, validating the effectiveness of the former for boosting energy efficiency, (2) G-Stationary dataflow has lower data access cost by keeping the generated $\mathbf{G}=\Hat{\mathbf{K}}^T \mathbf{V}$ stationary inner PEs which sequentially compute $\mathbf{Q}\mathbf{G}$ for enhancing data locality, and  
(3) Proposed down-forward accumulation dataflow instead reduces the more dominant systolic array cost by simplifying the PE design by advocating for input stationary when computing the matrix multiplications in the linear Taylor Attention.



\begin{table}[]
\centering
\caption{{Attention types and corresponding Pre/Post-Processors for typical linear attention Transformers.}}
\setlength{\tabcolsep}{0.3em}
\resizebox{\linewidth}{!}{
\begin{tabular}{|c|c|c|c|}
\hline
\textbf{Attention Types}                   & \textbf{Models}              & \textbf{Details}                                       & \textbf{Pre/Post-Processors} \\ \hline
Low-Rank                         & Linformer\cite{linformer}           & Reduce token dim. of $\mathbf{K}$/$\mathbf{V}$ & Exp. Div.           \\ \hline
\multirow{3}{*}{Kernel-Based} & Efficient Attention \cite{shen2021efficient} & $\phi()$= softmax()                             & Exp. Div.           \\ \cline{2-4} 
                                 & Performer \cite{choromanski2021rethinking}          & PORF    & Exp. Div. Add.      \\ \cline{2-4} 
                                 & Linear Transformer \cite{katharopoulos2020transformers}  & $\phi()$=elu() + 1                            & Exp. Div. Add.      \\ \hline
Taylor-Based                 & \textsc{ViTALiTy} (Ours)            & See Algorithm \ref{alg:vitality}         & Acc. Div. Add.      \\ \hline
\end{tabular}} \label{table:typical_transformers}
\vspace{-0.4cm}
\end{table}

\vspace{0.5em}

\textbf{Extension of \textsc{ViTALiTy} Accelerator.}
We summarize typical linear attention Transformers in Table \ref{table:typical_transformers}, where attention mechanisms are categorized into the low-rank based one \cite{linformer}, kernel-based ones \cite{shen2021efficient, choromanski2021rethinking, katharopoulos2020transformers}, and our proposed Taylor Attention.
We observe that 
our accelerator can be easily extended to accelerate diverse efficient Transformers with \textbf{dedicated pre/post-processors} for corresponding similarity ($\phi$) functions (e.g., softmax for \cite{linformer} and \cite{shen2021efficient}, PORF for \cite{choromanski2021rethinking}), and \textbf{generic systolic array} for matrix multiplications.    

%% file: main.bbl
\begin{thebibliography}{10}
\providecommand{\url}[1]{#1}
\csname url@samestyle\endcsname
\providecommand{\newblock}{\relax}
\providecommand{\bibinfo}[2]{#2}
\providecommand{\BIBentrySTDinterwordspacing}{\spaceskip=0pt\relax}
\providecommand{\BIBentryALTinterwordstretchfactor}{4}
\providecommand{\BIBentryALTinterwordspacing}{\spaceskip=\fontdimen2\font plus
\BIBentryALTinterwordstretchfactor\fontdimen3\font minus
  \fontdimen4\font\relax}
\providecommand{\BIBforeignlanguage}[2]{{%
\expandafter\ifx\csname l@#1\endcsname\relax
\typeout{** WARNING: IEEEtranS.bst: No hyphenation pattern has been}%
\typeout{** loaded for the language `#1'. Using the pattern for}%
\typeout{** the default language instead.}%
\else
\language=\csname l@#1\endcsname
\fi
#2}}
\providecommand{\BIBdecl}{\relax}
\BIBdecl

\bibitem{DC}
\BIBentryALTinterwordspacing
``Synopsys design compiler.'' [Online]. Available:
  \url{https://www.synopsys.com/implementation-and-signoff/rtl-synthesis-test/dc-ultra.html}
\BIBentrySTDinterwordspacing

\bibitem{beltagy2020longformer}
I.~Beltagy, M.~E. Peters, and A.~Cohan, ``Longformer: The long-document
  transformer,'' \emph{arXiv preprint arXiv:2004.05150}, 2020.

\bibitem{Longformer}
I.~Beltagy, M.~E. Peters, and A.~Cohan, ``Longformer: The long-document
  transformer,'' \emph{ArXiv}, vol. abs/2004.05150, 2020.

\bibitem{https://doi.org/10.48550/arxiv.2205.14756}
\BIBentryALTinterwordspacing
H.~Cai, C.~Gan, and S.~Han, ``Efficientvit: Enhanced linear attention for
  high-resolution low-computation visual recognition,'' 2022. [Online].
  Available: \url{https://arxiv.org/abs/2205.14756}
\BIBentrySTDinterwordspacing

\bibitem{carion2020end}
N.~Carion, F.~Massa, G.~Synnaeve, N.~Usunier, A.~Kirillov, and S.~Zagoruyko,
  ``End-to-end object detection with transformers,'' in \emph{European
  conference on computer vision}.\hskip 1em plus 0.5em minus 0.4em\relax
  Springer, 2020, pp. 213--229.

\bibitem{chen2021scatterbrain}
\BIBentryALTinterwordspacing
B.~Chen, T.~Dao, E.~Winsor, Z.~Song, A.~Rudra, and C.~R{\'e}, ``Scatterbrain:
  Unifying sparse and low-rank attention,'' in \emph{Advances in Neural
  Information Processing Systems}, A.~Beygelzimer, Y.~Dauphin, P.~Liang, and
  J.~W. Vaughan, Eds., 2021. [Online]. Available:
  \url{https://openreview.net/forum?id=SehIKudiIo1}
\BIBentrySTDinterwordspacing

\bibitem{chen2021crossvit}
C.-F.~R. Chen, Q.~Fan, and R.~Panda, ``Crossvit: Cross-attention multi-scale
  vision transformer for image classification,'' in \emph{Proceedings of the
  IEEE/CVF International Conference on Computer Vision}, 2021, pp. 357--366.

\bibitem{Chen_2021_ICCV}
H.~Chen, Z.~Luo, J.~Zhang, L.~Zhou, X.~Bai, Z.~Hu, C.-L. Tai, and L.~Quan,
  ``Learning to match features with seeded graph matching network,'' in
  \emph{Proceedings of the IEEE/CVF International Conference on Computer Vision
  (ICCV)}, October 2021, pp. 6301--6310.

\bibitem{chen2021chasing}
T.~Chen, Y.~Cheng, Z.~Gan, L.~Yuan, L.~Zhang, and Z.~Wang, ``Chasing sparsity
  in vision transformers: An end-to-end exploration,'' \emph{Advances in Neural
  Information Processing Systems}, vol.~34, pp. 19\,974--19\,988, 2021.

\bibitem{child2019generating}
R.~Child, S.~Gray, A.~Radford, and I.~Sutskever, ``Generating long sequences
  with sparse transformers,'' \emph{arXiv preprint arXiv:1904.10509}, 2019.

\bibitem{choromanski2021rethinking}
\BIBentryALTinterwordspacing
K.~M. Choromanski, V.~Likhosherstov, D.~Dohan, X.~Song, A.~Gane, T.~Sarlos,
  P.~Hawkins, J.~Q. Davis, A.~Mohiuddin, L.~Kaiser, D.~B. Belanger, L.~J.
  Colwell, and A.~Weller, ``Rethinking attention with performers,'' in
  \emph{International Conference on Learning Representations}, 2021. [Online].
  Available: \url{https://openreview.net/forum?id=Ua6zuk0WRH}
\BIBentrySTDinterwordspacing

\bibitem{correia2019adaptively}
G.~M. Correia, V.~Niculae, and A.~F. Martins, ``Adaptively sparse
  transformers,'' \emph{arXiv preprint arXiv:1909.00015}, 2019.

\bibitem{cui2019fine}
B.~Cui, Y.~Li, M.~Chen, and Z.~Zhang, ``Fine-tune bert with sparse
  self-attention mechanism,'' in \emph{Proceedings of the 2019 conference on
  empirical methods in natural language processing and the 9th international
  joint conference on natural language processing (EMNLP-IJCNLP)}, 2019, pp.
  3548--3553.

\bibitem{deng2009imagenet}
J.~Deng, W.~Dong, R.~Socher, L.-J. Li, K.~Li, and L.~Fei-Fei, ``Imagenet: A
  large-scale hierarchical image database,'' in \emph{2009 IEEE conference on
  computer vision and pattern recognition}.\hskip 1em plus 0.5em minus
  0.4em\relax Ieee, 2009, pp. 248--255.

\bibitem{devlin2018bert}
J.~Devlin, M.-W. Chang, K.~Lee, and K.~Toutanova, ``Bert: Pre-training of deep
  bidirectional transformers for language understanding,'' \emph{arXiv preprint
  arXiv:1810.04805}, 2018.

\bibitem{vit}
A.~Dosovitskiy, L.~Beyer, A.~Kolesnikov, D.~Weissenborn, X.~Zhai,
  T.~Unterthiner, M.~Dehghani, M.~Minderer, G.~Heigold, S.~Gelly, J.~Uszkoreit,
  and N.~Houlsby, ``An image is worth 16x16 words: Transformers for image
  recognition at scale,'' in \emph{International Conference on Learning
  Representations}, 2021.

\bibitem{germain2022visual}
\BIBentryALTinterwordspacing
H.~Germain, V.~Lepetit, and G.~Bourmaud, ``Visual correspondence
  hallucination,'' in \emph{International Conference on Learning
  Representations}, 2022. [Online]. Available:
  \url{https://openreview.net/forum?id=jaLDP8Hp_gc}
\BIBentrySTDinterwordspacing

\bibitem{nasvit}
C.~Gong, D.~Wang, M.~Li, X.~Chen, Z.~Yan, Y.~Tian, qiang liu, and V.~Chandra,
  ``{NASV}it: Neural architecture search for efficient vision transformers with
  gradient conflict aware supernet training,'' in \emph{International
  Conference on Learning Representations}, 2022.

\bibitem{pixel3}
{Google LLC.}, ``{Pixel 3},'' \url{https://g.co/kgs/pVRc1Y}, accessed
  2020-09-01.

\bibitem{Graham_2021_ICCV}
B.~Graham, A.~El-Nouby, H.~Touvron, P.~Stock, A.~Joulin, H.~Jegou, and
  M.~Douze, ``Levit: A vision transformer in convnet's clothing for faster
  inference,'' in \emph{Proceedings of the IEEE/CVF International Conference on
  Computer Vision (ICCV)}, October 2021, pp. 12\,259--12\,269.

\bibitem{ham20203}
T.~J. Ham, S.~J. Jung, S.~Kim, Y.~H. Oh, Y.~Park, Y.~Song, J.-H. Park, S.~Lee,
  K.~Park, J.~W. Lee \emph{et~al.}, ``A\^{} 3: Accelerating attention
  mechanisms in neural networks with approximation,'' in \emph{2020 IEEE
  International Symposium on High Performance Computer Architecture
  (HPCA)}.\hskip 1em plus 0.5em minus 0.4em\relax IEEE, 2020, pp. 328--341.

\bibitem{ham2021elsa}
T.~J. Ham, Y.~Lee, S.~H. Seo, S.~Kim, H.~Choi, S.~J. Jung, and J.~W. Lee,
  ``Elsa: Hardware-software co-design for efficient, lightweight self-attention
  mechanism in neural networks,'' in \emph{2021 ACM/IEEE 48th Annual
  International Symposium on Computer Architecture (ISCA)}.\hskip 1em plus
  0.5em minus 0.4em\relax IEEE, 2021, pp. 692--705.

\bibitem{heo2021rethinking}
B.~Heo, S.~Yun, D.~Han, S.~Chun, J.~Choe, and S.~J. Oh, ``Rethinking spatial
  dimensions of vision transformers,'' in \emph{Proceedings of the IEEE/CVF
  International Conference on Computer Vision}, 2021, pp. 11\,936--11\,945.

\bibitem{katharopoulos2020transformers}
A.~Katharopoulos, A.~Vyas, N.~Pappas, and F.~Fleuret, ``Transformers are rnns:
  Fast autoregressive transformers with linear attention,'' in
  \emph{International Conference on Machine Learning}.\hskip 1em plus 0.5em
  minus 0.4em\relax PMLR, 2020, pp. 5156--5165.

\bibitem{Kitaev2020Reformer}
\BIBentryALTinterwordspacing
N.~Kitaev, L.~Kaiser, and A.~Levskaya, ``Reformer: The efficient transformer,''
  in \emph{International Conference on Learning Representations}, 2020.
  [Online]. Available: \url{https://openreview.net/forum?id=rkgNKkHtvB}
\BIBentrySTDinterwordspacing

\bibitem{liu2021Swin}
Z.~Liu, Y.~Lin, Y.~Cao, H.~Hu, Y.~Wei, Z.~Zhang, S.~Lin, and B.~Guo, ``Swin
  transformer: Hierarchical vision transformer using shifted windows,'' in
  \emph{Proceedings of the IEEE/CVF International Conference on Computer
  Vision}, 2021, pp. 10\,012--10\,022.

\bibitem{vit_quantization}
Z.~Liu, Y.~Wang, K.~Han, W.~Zhang, S.~Ma, and W.~Gao, ``Post-training
  quantization for vision transformer,'' in \emph{Advances in Neural
  Information Processing Systems}, M.~Ranzato, A.~Beygelzimer, Y.~Dauphin,
  P.~Liang, and J.~W. Vaughan, Eds., vol.~34.\hskip 1em plus 0.5em minus
  0.4em\relax Curran Associates, Inc., 2021, pp. 28\,092--28\,103.

\bibitem{sanger}
L.~Lu, Y.~Jin, H.~Bi, Z.~Luo, P.~Li, T.~Wang, and Y.~Liang, ``Sanger: A
  co-design framework for enabling sparse attention using reconfigurable
  architecture,'' ser. MICRO '21.\hskip 1em plus 0.5em minus 0.4em\relax New
  York, NY, USA: Association for Computing Machinery, 2021, p. 977–991.

\bibitem{lu2021sanger}
L.~Lu, Y.~Jin, H.~Bi, Z.~Luo, P.~Li, T.~Wang, and Y.~Liang, ``Sanger: A
  co-design framework for enabling sparse attention using reconfigurable
  architecture,'' in \emph{MICRO-54: 54th Annual IEEE/ACM International
  Symposium on Microarchitecture}, 2021, pp. 977--991.

\bibitem{mehta2021mobilevit}
S.~Mehta and M.~Rastegari, ``Mobilevit: light-weight, general-purpose, and
  mobile-friendly vision transformer,'' \emph{arXiv preprint arXiv:2110.02178},
  2021.

\bibitem{tx2}
{NVIDIA Inc.}, ``{NVIDIA Jetson TX2},''
  \url{https://www.nvidia.com/en-us/autonomous-machines/embedded-systems/jetson-tx2/},
  accessed 2020-09-01.

\bibitem{2080ti}
{NVIDIA LLC.}, ``{GeForce RTX 2080 TI Graphics Card | NVIDIA},'' 2021,
  \url{https://www.nvidia.com/en-me/geforce/graphics-cards/rtx-2080-ti/},
  accessed 2020-09-01.

\bibitem{qu2022dota}
Z.~Qu, L.~Liu, F.~Tu, Z.~Chen, Y.~Ding, and Y.~Xie, ``Dota: detect and omit
  weak attentions for scalable transformer acceleration,'' in \emph{Proceedings
  of the 27th ACM International Conference on Architectural Support for
  Programming Languages and Operating Systems}, 2022, pp. 14--26.

\bibitem{radford2018improving}
A.~Radford, K.~Narasimhan, T.~Salimans, and I.~Sutskever, ``Improving language
  understanding by generative pre-training,'' 2018.

\bibitem{SALO}
G.~Shen, J.~Zhao, Q.~Chen, J.~Leng, C.~Li, and M.~Guo, ``Salo: an efficient
  spatial accelerator enabling hybrid sparse attention mechanisms for long
  sequences,'' \emph{Proceedings of the 59th ACM/IEEE Design Automation
  Conference}, 2022.

\bibitem{shen2021efficient}
Z.~Shen, M.~Zhang, H.~Zhao, S.~Yi, and H.~Li, ``Efficient attention: Attention
  with linear complexities,'' in \emph{Proceedings of the IEEE/CVF winter
  conference on applications of computer vision}, 2021, pp. 3531--3539.

\bibitem{tang2022quadtree}
\BIBentryALTinterwordspacing
S.~Tang, J.~Zhang, S.~Zhu, and P.~Tan, ``Quadtree attention for vision
  transformers,'' in \emph{International Conference on Learning
  Representations}, 2022. [Online]. Available:
  \url{https://openreview.net/forum?id=fR-EnKWL_Zb}
\BIBentrySTDinterwordspacing

\bibitem{pmlr-v139-touvron21a}
H.~Touvron, M.~Cord, M.~Douze, F.~Massa, A.~Sablayrolles, and H.~Jegou,
  ``Training data-efficient image transformers \& distillation through
  attention,'' in \emph{International Conference on Machine Learning}, vol.
  139, July 2021, pp. 10\,347--10\,357.

\bibitem{vaswani2017attention}
A.~Vaswani, N.~Shazeer, N.~Parmar, J.~Uszkoreit, L.~Jones, A.~N. Gomez,
  {\L}.~Kaiser, and I.~Polosukhin, ``Attention is all you need,''
  \emph{Advances in neural information processing systems}, vol.~30, 2017.

\bibitem{wang2021spatten}
H.~Wang, Z.~Zhang, and S.~Han, ``Spatten: Efficient sparse attention
  architecture with cascade token and head pruning,'' in \emph{2021 IEEE
  International Symposium on High-Performance Computer Architecture
  (HPCA)}.\hskip 1em plus 0.5em minus 0.4em\relax IEEE, 2021, pp. 97--110.

\bibitem{linformer}
\BIBentryALTinterwordspacing
S.~Wang, B.~Z. Li, M.~Khabsa, H.~Fang, and H.~Ma, ``Linformer: Self-attention
  with linear complexity,'' 2020. [Online]. Available:
  \url{https://arxiv.org/abs/2006.04768}
\BIBentrySTDinterwordspacing

\bibitem{wang2021pyramid}
W.~Wang, E.~Xie, X.~Li, D.-P. Fan, K.~Song, D.~Liang, T.~Lu, P.~Luo, and
  L.~Shao, ``Pyramid vision transformer: A versatile backbone for dense
  prediction without convolutions,'' in \emph{Proceedings of the IEEE/CVF
  International Conference on Computer Vision}, 2021, pp. 568--578.

\bibitem{rw2019timm}
R.~Wightman, ``Pytorch image models,''
  \url{https://github.com/rwightman/pytorch-image-models}, 2019.

\bibitem{wu2021cvt}
H.~Wu, B.~Xiao, N.~Codella, M.~Liu, X.~Dai, L.~Yuan, and L.~Zhang, ``Cvt:
  Introducing convolutions to vision transformers,'' 2021.

\bibitem{yang2021focal}
J.~Yang, C.~Li, P.~Zhang, X.~Dai, B.~Xiao, L.~Yuan, and J.~Gao, ``Focal
  self-attention for local-global interactions in vision transformers,''
  \emph{arXiv preprint arXiv:2107.00641}, 2021.

\bibitem{ye2019cross}
L.~Ye, M.~Rochan, Z.~Liu, and Y.~Wang, ``Cross-modal self-attention network for
  referring image segmentation,'' in \emph{Proceedings of the IEEE/CVF
  Conference on Computer Vision and Pattern Recognition}, 2019, pp.
  10\,502--10\,511.

\bibitem{yu2022unified}
S.~Yu, T.~Chen, J.~Shen, H.~Yuan, J.~Tan, S.~Yang, J.~Liu, and Z.~Wang,
  ``Unified visual transformer compression,'' \emph{arXiv preprint
  arXiv:2203.08243}, 2022.

\bibitem{zaheer2020big}
M.~Zaheer, G.~Guruganesh, K.~A. Dubey, J.~Ainslie, C.~Alberti, S.~Ontanon,
  P.~Pham, A.~Ravula, Q.~Wang, L.~Yang \emph{et~al.}, ``Big bird: Transformers
  for longer sequences,'' \emph{Advances in Neural Information Processing
  Systems}, vol.~33, pp. 17\,283--17\,297, 2020.

\bibitem{zhao2019explicit}
G.~Zhao, J.~Lin, Z.~Zhang, X.~Ren, Q.~Su, and X.~Sun, ``Explicit sparse
  transformer: Concentrated attention through explicit selection,'' \emph{arXiv
  preprint arXiv:1912.11637}, 2019.

\bibitem{zhu2021visual}
M.~Zhu, K.~Han, Y.~Tang, and Y.~Wang, ``Visual transformer pruning,''
  \emph{arXiv e-prints}, pp. arXiv--2104, 2021.

\bibitem{zhu2020deformable}
X.~Zhu, W.~Su, L.~Lu, B.~Li, X.~Wang, and J.~Dai, ``Deformable detr: Deformable
  transformers for end-to-end object detection,'' \emph{arXiv preprint
  arXiv:2010.04159}, 2020.

\end{thebibliography}
